\documentclass{IEEEtran}

\usepackage{amssymb, amsmath, verbatim, amsthm,url, multirow,fullpage,mathtools, graphicx, subcaption, csvsimple, outlines}
\usepackage[all]{xy}
\usepackage{booktabs}

\usepackage{soul}
\usepackage[colorinlistoftodos,textsize=footnotesize]{todonotes}

\setstcolor{red}

\newcommand{\cM}{\mathcal{M}}
\newcommand{\cI}{\mathcal{I}}
\newcommand{\cR}{\mathcal{R}}
\newcommand{\cL}{\mathcal{L}}
\newcommand{\cD}{\mathcal{D}}
\newcommand{\cU}{\mathcal{U}}
\newcommand{\cO}{\mathcal{O}}

\newcommand{\R}{\mathbb{R}}

\newcommand{\bI}{\mathbb{I}}
\newcommand{\D}{\partial}

\def\ba #1\ea{\begin{align} #1 \end{align}}
\def\bas #1\eas{\begin{align*} #1 \end{align*}}
\def\bml #1\eml{\begin{multline} #1 \end{multline}}
\def\bmls #1\emls{\begin{multline*} #1 \end{multline*}}

\newtheorem{thm}{Theorem}[section]

\theoremstyle{remark}

\theoremstyle{definition}

\usepackage{authblk}

\title{Geometric instability of out of distribution data across autoencoder architecture}

\author[1]{Susama Agarwala}
\author[2]{Benjamin Dees}
\author[1]{Corey Lowman}

\affil[1]{Johns Hopkins University Applied Physics Lab}
\affil[2]{Johns Hopking University, Mathematics Department}

\date{\today}                     
\begin{document}

\maketitle 

\begin{abstract}
We study the map learned by a family of autoencoders trained on MNIST, and evaluated on ten different data sets created by the random selection of pixel values according to ten different distributions. Specifically, we study the eigenvalues of the Jacobians defined by the weight matrices of the autoencoder at each training and evaluation point. For high enough latent dimension, we find that each autoencoder reconstructs all the evaluation data sets as similar \emph{generalized characters}, but that this reconstructed \emph{generalized character} changes across autoencoder. Eigenvalue analysis shows that even when the reconstructed image appears to be an MNIST character for all out of distribution data sets, not all have latent representations that are close to the latent representation of MNIST characters. All told, the eigenvalue analysis demonstrated a great deal of geometric instability of the autoencoder both as a function on out of distribution inputs, and across architectures on the same set of inputs.\end{abstract}

The distributions of training, test and validation data is often different than the distribution of the data on which a neural network is deployed. This problem of distribution shift causes neural networks to fail silently, with potentially serious consequences \cite{AmazonRekognition,wilson2019predictive}. The research community has tried to address this in many ways with different levels of success. For instance, generative models \cite{nalisnick2018deep, havtorn2021hierarchical} effectively learn the distributions of the probability distribution of the training data set, but they do not address the fact that one often has little control over the deployment data distribution. To address this problem, the research community has come up with a wide family of out-of-distribution detection algorithms \cite{hendrycks2016baseline, liang2017enhancing, devries2018learning, krueger2021out}, with a variety of strengths and weaknesses. 

However, to our knowledge, there has not been any concerted effort to understand the geometric properties of the feature map that transforms inputs to a latent representation. While this map is trained on the training, test and validation data distributions, it is a function on a much larger space of possible inputs. In this paper, we consider an autoencoder as a \emph{function} from input to reconstruction space, such that one can consider its Jacobian matrix at any input point. The eigenvalues and eigenvectors of these Jacobians contain a wealth of information about the geometric properties of the feature map. In \cite{gag}, the authors studied the geometry of this function on the training data set (MNIST). In this paper, we consider the geometry on points very far out of distribution (specifically, on 10 data sets where pixel brightness is drawn from 10 different distributions). 

As autoencoders are supposed to be a method for learning the manifold structure of a data set, they are frequently used as means of out of distribution detection \cite{Zhao17, liang2018enhancing}. However, this method of out of distribution detection fails due to the well documented phenomenon of out of distribution reconstruction \cite{nalisnick2018deep, yoon2021autoencoding, xiao2020likelihood, denouden2018improving}. We consider the geometric stability of the learned feature map on points so far from the training data set that out of distribution reconstruction is not a worry, and characterize how these geometric properties are different from those on the training set. We believe this line of inquiry will lead to a foundational understanding of what an autoencoder is learning about the geometry of the data manifold as well as the relation to the input space, and lead to a new set of geometric tools for out of distribution detection.

Concretely, we evaluate a family of autoencoders with identical architecture except for the dimension of the latent space for various seeds. Then we evaluate 10 data sets, each with pixel values drawn from a different random distribution on these trained networks. We find that on data sets far out of distribution, at low latent dimension, the reconstruction often appears to be a reconstructed image (under visual inspection) from the training data set. However, the autoencoder will map any given random distribution of pixels to a variety of different training images, and this set will change from seed to seed. Surprisingly, however, the geometric data suggests that even though the reconstruction images appear to be close to the reconstructed images of the training data set, the latent representation of the training data and the out of distribution data is quite different. 
Furthermore, we find that for higher latent dimensions, the autoencoder consistently maps each random pixel distribution data set to the same class of \emph{generalized character}, but that this character is both seed and architecture dependent. We find that the geometries of the learned feature functions cluster by mean value of the pixel distribution. More surprisingly, we learn that for higher latent dimension, the trained autoencoder is not aware of the orientation of the bases of the input space.

\section{Autoencoders and learned geometry}
An autoencoder consists of a pair of neural networks trained in tandem, an encoder that maps points from a high dimensional input space, $\cI \simeq \R^N$ to a lower dimensional latent space $\cL \simeq \R^d$, and a decoder that maps the points in the latent space back to a high dimensional reconstruction space (isomorphic to the input space), $\cR \simeq \R^N$, with $d << N$. Let $f_{enc}$ indicate the encoder network, and $f_{dec}$ the decoder. For any input point $x \in \cI$, let $y = f_{dec} \circ f_{enc}(x) \in \cR$ be the reconstructed point and $z = f_{enc}(x) \in \cL$ the latent representation of said point. By construction, the image of the autoencoder $f_{dec} \circ f_{enc}: \cI \rightarrow \cR$ is at most a $d$ dimensional subspace of $\cR$. For the purposes of this paper, the autoencoders in question are trained to minimize the reconstruction loss, i.e. to minimize the average Euclidean distance between the input and the corresponding reconstruction point. 

\subsection{Neural Networks and their Jacobians}
In this paper, we think of any neural network as the composition of the functions represented by the weight matrices, bias vectors, and the activation functions in each layer.  In other words, if a neural network has an $N$ dimensional input space and an $M$ dimensional output, we may represent it as a multi-variable function, $NN: \R^N \rightarrow \R^M$. The Jacobian matrix of a multi-variable function is the matrix of gradient vectors of each of the components of the function, and is represented as a $M \times N$ matrix with function valued entries. In general, this Jacobian matrix cannot be represented or computed in function form, however, one can always talk about its the value of the Jacobian matrix at any given input point $x$, $J_{NN}(x)$. This matrix represents a linear transformation from the tangent space the input space at $x$ to the tangent space of the output space at the output point $y = NN(x)$. One may always study the singular value decomposition of $J_{NN}(x)$ to find an orthonormal basis of the tangent space $T_x(\R^N)$ and understand how it transforms in terms of an orthonormal basis of $T_y(\R^M)$. Note that the Jacobean matrix is only defined where the function is  differentiable. In the case of neural networks with ReLU activation functions, as is the case for the autoencoders considered in this paper, the functions are piece wise linear. Therefore, they are not differentiable on a space of measure $0$ in $\R^N$. In other words, the matrix $J_{NN}(x)$ is not defined with probability $0$. 

In the special case of $N= M$, as is the case for autoencoders, one may also consider the eigenvalues and eigenvectors of the Jacobian matrix, which quantifies the directions and amounts in which the map $J_{NN}(x)$ dilates or contracts the tangent space $T_x(\R^N)$. This is fundamentally different from a singular value decomposition of $J_{NN}(x)$ as it identifies a common basis for the two tangent spaces, $T_x(\R^N)$ and $T_y(\R^N)$, given by the eigenvectors. In other words, the eigenvalues gives insight into the local deformations of the input space as the function learned by the autoencoder maps it to the reconstruction space.  

The Jacobian matrix of the autoencoder is computed by $J_\cI := D \big(f_{dec} \circ f_{enc}\big) (x) = D f_{dec}(z) \cdot D f_{enc}(x)$. See  Appendix \ref{app:Jacobean} for details. For any matrix square matrix $A$, let $\vec{\lambda}_A$ be the vector of norm decreasing eigenvalues of $A$. Note that the image of the autoencoder in the reconstruction space is at most $d$ dimensional, for $\cL = \R^d$. Therefore, for any point $x \in \cI$, the vector $\vec{\lambda}_{J_\cI(x)}$ will have at most $d$ non-zero entries, each indicating the ammount of stretching or warping that the corresponding eigenspace of the data point (as a subspace of $T_x \cI$) undergoes. In particular, the eigenspace of $0$ corresponds to the kernel of $J_\cI(x)$. Geometrically, these are the directions of the input space that are collapsed to the origin in the image of the autoencoder. 


Calculating the eigenvalues of $J_\cI(x)$ is computationally expensive, as algorithms for calculating eigenvalues of an $n\times n$ matrix are $O(n^3)$ \cite{QRalg}. Therefore, it is easier to calculate the eigenvalues of the Jacobian of a related function, $J_\cL:= D \big( f_{enc} \circ f_{dec} \big) (x) = D f_{enc}(y) \cdot f_{dec}(z)$ which represents the deformations of the latent space under the action of the autoencoder composed in the opposite order. In general, there is no reason to expect that these two Jacobians should have any similarity in their eigenvalues. However, in 
 \cite{gag}, the authors show that when the the reconstruction error is $0$,the eigenvalues of $J_\cL$ are similar to the eigenvalues of $J_\cI$. \begin{thm} \label{res:evalssame}
If $f_{dec}\circ f_{enc}(x)=x$, then for $z = f_{enc}(x)$, the $d$ nonzero eigenvalues of $J_\cI(x)$ are the same as the eigenvalues of $J_\cL(z)$. 
\end{thm}  The proof of this theorem is given in \cite{gag}, but is included in Appendix \ref{app:geometry} for completeness. Empirically, we see that when the reconstruction error is small, these values are close, but not identical. This property will not hold when the reconstruction error is large (see Figure \ref{fig:eigennorms}). 

\subsection{Probing the data manifold structure}

The Jacobians $J_\cI(x)$ and $J_\cL(z)$ give insight into how well an autoencoder is learning the data manifold underlying the training data. 

The data manifold conjecture simply states that, for any given data set of related objects $\cD$, the data actually lies noisily around some low ($\delta$) dimensional manifold $M_\cD$ \cite{bengio2013representation, alain2014regularized}. If a neural network learns a good representation of the data, then it learns the structure of $M_\cD$. In topological terms, learning the structure of $M_\cD$ consists of learning a diffeomorphism $\phi_\cU$ from an open set $\cU \subset M_\cD$ to $\R^\delta$. Ideally, if $\delta > d$, then  the learned function restricted to the data manifold, $f_{enc}(M_\cD)$ is the composition of $\phi_\cU$ with a projection onto the latent space: $\pi: \R^\delta \rightarrow \R^d$. If the latent space has greater dimension than the data manifold ($\delta \leq d$) then $f_{enc}(M_\cD)$ is the composition of $\phi_\cU$ with an embedding into the latent space: $\pi: \R^\delta \hookrightarrow \R^d$. The function $f_{dec}(\cL)$ is the inversion of this function. The commutative diagram for this interaction is shown in display \eqref{disp:traindiagram}.

The function $f_{enc}$ is not just defined on the points near $M_\cD$, but on all of $\cI$. Little is understood about the geometry of the neural network $f_{enc}$ on points far away from the data manifold. Let $\cM \subset \cL$ be the image of $M_\cD$ under $f_{enc}$. Since the input space has much greater dimension than the latent space, $N >>d$, $f_{enc}$ has a large kernel. That is, many out of distribution points will map onto $\cM$. Then $f_{dec}$ maps all these points to the the same reconstruction point.

\begin{thm} \label{res:JLbound}
Let $x \in \cI$ be an input point with latent representation, $z' = f_{enc}(x)$, is close to a point $z \in \cM$ in the image of the data manifold in latent space.  Then \bmls ||\vec{\lambda}_{J_\cL(z)} - \vec{\lambda}_{J_\cL(z')}||^2 \leq ||J_\cL(z') - J_\cL(z)||^2_F \\  = (\cO(\|z'-z\|^2) )^2\;.\emls 
\end{thm}

Therefore, the distance between the eigenvalue vectors of $J_{\cL}(z')$ and $J_{\cL}(z)$ gives a lower bound on the Frobenius norm of difference of the Jacobean matrices. Namely, even if an out of distribution point is reconstructed to an element of $f_{dec} \circ f_{enc}(M_\cD) \subset \cR$, the eigenvalues of $J_{\cL}(z')$ gives information about its latent representation.

\subsection{In distribution eigenvalue behavior \label{sec:indisttheory}}

Let $\cD_{train}$ be the training data set, and $M_{\cD_{train}}$ the corresponding $\delta_{train}$ dimensional data manifold. Minimizing the reconstruction error ensures the the auto-encoder is a good zeroth order model for a sample of points in and around $M_{\cD_{train}}$. Studying the eigenvalues of the Jacobian $J_\cI(x)$ gives information about how well the autoencoder performs as a first order approximation of data manifold,locally around $x$. For instance, suppose $d \leq \delta_{train}$. If the top $d$ eignevalues of $J_\cI(x)$ are one, then the auto-encoder has projected exactly onto the manifold $M_{\cD_{train}}$ at each input point $x$. If the arguments of the eigenvalues of $J_\cI(x)$ are $0$, then there is little rotation locally in passing from each input to reconstruction point. If the product of the top $d$ eigenvalues of $J_\cI(x) > 0$, then there is  no local orientation reversing behavior the input point. 

Furthermore, for $\delta_{train} < d$, if the autoencoder projects onto the tangent of the data manifold, then one would expect to see $d-\delta_{train}$ $0$ eigenvalues for each $J_{\cI}(x)$. At the the other extreme, if the autoencoder projects onto all of $\cL$, instead of onto a $\delta_{train}$ dimensional submanifold, then one would expect to see the top $d$  eigenvalues of $J_\cI(x)$ and $J_\cL(z)$ remain at $1$ for all $d$. 

\subsection{Out of distribution eigenvalue behavior \label{sec:outdisttheory}}
Let $\cD$ be a fundamentally different data set than $\cD_{train}$. For $x \in \cD$, one cannot use the norms of the eigenvalues of $J_\cI(x)$ alone to understand whether or not $f_{enc}$ is projecting $T_x(\cI)$ onto the tangent space of point on $M_{\cD_{train}}$. To see why, consider the points on the inside and the outside of a $d$ dimensional unit sphere, $S^d$ embedded inside $R^N$. Consider the points that lie on the rays emanating from the origin through $S^d$ in $\R^N$. For points lying close to the origin, any projection onto the surface of the sphere will have the properties that a small perturbation in the domain will result in a large perturbation on the sphere. I.e. the eigenvalues will be large. For projections of the points on the rays on the outside of the sphere, however, a large perturbation in the domain will correspond to a small perturbation on the surface. I.e. the eigenvalues will be small. 

Furthermore, the average size of the argument of the eigenvalues and the sign of the product of the eigenvalues of $J_\cI(x)$ has structure if the function $f_{enc}(x)$ factors through a projection onto $M_{\cD_{train}}$. 

\begin{thm} \label{res:factor} Let $x \in \cD$ be far away from the data manifold $M_{\cD_{train}}$. If the function $f_{enc}(x)$ factors through a parallel transport onto $M_{\cD_{train}}$, the sign of the product of the non-zero eigenvalues will be positive (the Jacobian does not reverse orientation from $T_x\cI$ to $T_y\cR$), and the arguments of the eigenvalues will be zero (the Jacobian does induce a rotation from $T_x\cI$ to $T_y\cR$). \end{thm}

Finally, it is worth noting that an autoencoder trained on $M_{\cD_{train}}$ will not capture information on the geometry of $M_\cD$. For instance, there is no reason for an autoencoder $f_{dec}\circ f_{enc}$ that has learned the geometry of $M_{\cD_{train}}$ (as outlined in Section \ref{sec:indisttheory}) will have captured any information about the geometry of a distant evaluation manifold $M_{\cD}$. For instance, the appearance of zero eigenvalues after latent dimension $d$ in the Jacobean $J_\cI(x)$, for $x \in \cD$ should not indicate that $M_{\cD}$ has dimension $d$. Rather, it indicates that the trained autoencoder is keeping track of fewer than $d$ of the dimensions of the input space near $x$.

\subsection{Observed behavior}

In this paper, we observe that for an $x \in M_{\cD_{train}}$, the autoencoder displays several properties one would expect if the map $f_{dec} \circ f_{enc}(x)$ induced a projection onto $T_yM_{\cD_{train}}$. Namely, the norm of the eigenvalues are, on average, less than one, but remain relatively close to it for both $J_\cI(x)$ and $J_\cL(z)$. Furthermore, as the latent dimension increases, the distribution of eigenvalues develop increasingly long left tails on a log scale, see Figure \ref{fig:lognormbox}. That is, the proportion of eigenvalues with small norms increases, though not on as quickly as one would expect for a projection. Furthermore, the arguments of the eigenvalues are small, as are the proportion of points with locally orientation reversing behavior. This is consistent with work studying autoencoders as an iterative system, where the output of a trained autoencoder is evaluated again by the same autoencoder, where one finds that the outcome is not the same, but does converge \cite{memorization}. Furthermore, this behavior is consistent across autoencoder architecture.

However, this behavior is not observed for data sets that are far out of distribution. For instance, we note that, for $J_\cL$, the arithmetic mean of the eigenvalues of the out of distribution points is frequently quite different from the arithmetic mean of the eigenvalues of the in distribution points (Figure \ref{fig:eigenmeans}), indicating that Frobenius norms of the two matrices are large (by Theorem \ref{res:JLbound}), even when, visually, the reconstruction image appears to be in $M_{\cD_{train}}$. In other words, even when the reconstructed image appears to be in the reconstruction of the training manifold, the latent representation of the out of domain point can be quite different from the latent representations of the training data.

Furthermore, when the latent dimension is large, about half of the out of distribution data points show orientation reversing behavior by $J_\cI(x)$ and the eigenvalues have a larger range of arguments than in the training data, indicating that the map $f_{enc}$ does not factor through a parallel transport onto $M_{\cD_{train}}$. Since approximately $50\%$ of the points display orientation reversing data, the autoencoder has not learned anything about the orientation of the tangent space, $T_x\cI$, far out of distribution.

\section{Experiment}

For this paper, we trained nineteen autoencoders on the MNIST data set, each with different latent dimensions, but otherwise identical architectures. We then evaluated these nineteen neural networks on 10 different out of distribution data sets, each consisting of 70,000 draws from different random distributions on the pixel values. Henceforth, we refer to the full MNIST data set as $\cD_{train}$, even though the training was done on the standard 60,000 training points of the 70,000 point sample.

We observe several notable empirical results. First, at low latent dimension, the autoencoders reconstructs the out of domain data as MNIST images, though inconsistently. As the latent dimension increases, the autoencoders stop reconstructing the data sets at MNIST images, but rather reconstruct them as \emph{generalized characters}. It is interesting to note, however, that while the autoencoder seems to have learned some general features of written characters from the MNIST data set, there is no consistency of the reconstructions across either architectures or seeds, indicating the instability of network performance on out of domain data. 

As further evidence of the instability of the autoencoders, we note variation in the arithmetic and geometric means of the eigenvalues, and that the standard deviation of the log normal of the eigenvalues is, in general greater for the out of distribution data than for the training data (see Figure \ref{fig:stdev}). We consider the log normal of the eigenvalues rather than just the eigenvalues themselves in order to capture the expectation that as latent dimension increases past the intrinsic dimension of $M_{\cD_{train}}$. While one expects the emergence of more zero (or very small in norm) eigenvalues, which we do not observe. The increased variation within each autoencoder indicates a lack of consistency of the eigenvalues of $J_\cI(x)$ on each out of domain point $x$. The increased variation across autoencoders is further evidence of the instability of network performance on out of domain data. 

Finally, inspite of the instability or reconstructions across architectures, certain patterns still emerge. Namely, the arithmetic and geometric means of the eigenvalues cluster according to the means of the distributions from which they are drawn. In other words, while the autoencoders do not geometrically encode much data about out of distribution points, they do learn information about the mean pixel values.

\subsection{Data and architecture}
For this paper, we trained nineteen unregularized autoencoders on the MNIST data set on two seeds. While we are aware that there are many regularization processes that will improve the accuracy and/ or generalization abilities of out autoencoder, we intentionally work with unregularized networks as these have the easiest to interpret geometric properties. Each autoencoder has four identical layers in the encoder and decoder differeing only by the dimension of the latent dimension: the encoder half consists of layers mapping between spaces of dimension $(784, 128, 64, 32, d)$, and the decoder half consists of layers between dimensions $(d, 32, 64, 128, 784)$, with $d \in \{2, \ldots, 19\}$. 

The MNIST images are 28x28 arrays flattened to a vector for input. Thus the input and reconstruction spaces are $\cI = [0,1]^{784} = \cR$. Furthermore, the decoder has a final $\tanh$ function to renormalize the output to a vector with values in the interval $[0,1]$. The activation function for all layers is ReLU, making the autoencoder represent a piece wise linear function, and is trained for 300 epochs, with seeds $0$ and $1$. 

After training, we evaluate 10 diferent samples of 70,000 datapoints drawn from 10 different out of distribution data sets, each defined by a random distribution on $\cI$, with different means: truncated normal between 0 and 1, (mean 0.5), uniform over $[0,1]$ (mean .5), 4 Bernoulli distribution with means $0.13$,  $0.25$, $0.5$, $0.75$, and $0.87$, and three Beta distributions one with $\alpha = .8$, $\beta = 5$, mean .14, another with $\alpha = .5$, $\beta = .5$, mean .5, and the final with $\alpha = 5$, $\beta = .8$, mean = .86. We refer to the data sets as $\cD_{norm}$, $\cD_{unif}$, $\cD_p$ for the 4 Bernoulli distributions (where $p \in \{.13, .25, .5, .74, .87\}$ and $\cD_{\alpha, \beta}$ for the three Beta distributions (where $(\alpha, \beta) \in \{(.8, 5), (.5, .5), (5, .8)\}$). 

We note that the mean value for the MNIST data set is .13, with most of the the values being 0 (black), and a small percentage being near 1 (white). While the location of the bright pixels is clearly important for the classifier, we chose $\cD_{.13}$ and $\cD_{.8, 5}$ to mimic this behavior. The data sets $\cD_{.87}$ and $\cD_{5, .8}$ are chosen to reverse this behavior. All other distributions are chosen to interpolate between these two extremes.  

\subsection{Geometry of out of distribution data}

As we draw the out of distribution data from well defined distributions, we may characterize this data in terms of several data manifolds. Most simply, we may consider each data set as being noisily distributed around a $0$ dimensional manifold, the point where each pixel has the mean brightness. In this manner, we have $10$ different $0$ dimensional data manifolds, indicated as above as $M_{\cD_p}$, $M_{\cD_{\alpha, \beta}}$, $M_{\cD_{norm}}$ and $M_{\cD_{unif}}$. Note that given the similarities of the means, from this perspective, the manifolds $M_{\cD_{.13}}$ and $M_{\cD_{.8, 5}}$ are close; $M_{\cD_{norm}}$,  $M_{\cD_{unif}}$ and $M_{\cD_{.8, 5}}$, are the same; and $M_{\cD_{.87}}$ and $M_{\cD_{5,.8}}$ are close together. However, the corresponding data sets \emph{are} different, as the sampling around the point corresponding to the manifold is different in each case. 

We may also consider these points to be lying on a larger data manifold parameterized by the properties of the distributions. Allowing $p$ to vary from 0 to 1, we obtain a 1 dimensional data manifold $M_{Bernoulli}$, where each $\cD_p$ is a sample of points distributed noisily (according to the appropriate Bernoulli distribution) about the point $p \in M_{Bernoulli}$. Similarly, there is a two dimensional $M_{Beta}$ parametrized by $0 < \alpha, \; \beta$ where each $\cD_{\alpha, \beta}$ is a sample of points distributed noisily according to the appropriate Beta distribution. Finally, $\cD_{norm}$ and $\cD_{unif}$ come from tow points on the two dimensional manifold $M_{norm}$, parametrized by $\mu$ and $\sigma$ with data is distributed around it according to the truncated normal distribution with mean $0 < \mu < 1$ and $0 <\sigma <\infty$: $\cD_{norm}$ samples around the point $ \mu = .5$, $\sigma = 1$ and $\cD_{unif}$ around $ \mu = .5$, $\sigma = \infty$. 

Note that, as discussed in Section \ref{sec:outdisttheory}, the eigenvalues of the $J_\cI(x)$ do not capture the intrinsic dimension of the out of distribution data. Namely, one does not observe the emergence of $0$ (or small normed) eigenvalues at low latent dimension, as one would expect if the autoencoder had learned the intrinsic dimension (see Figure \ref{fig:lognormbox}). 
 
\subsection{Reconstruction images}
Looking at these 10 data sets, evaluated on the 19 different autoencoders, we find that for each autoencoder with large enough latent dimension, the reconstructed images can qualitatively be placed into two groups, those with mean less than .5, and those with means at least .5 (see Tables \ref{tbl:reconstructions} and \ref{tbl:reconstructions1}). Examining the behavior of the eigenvalues, we get a finer granularity of three mean based clusters: those with mean less than .5, those with mean .5 and those with mean greater than .5 (see Figure \ref{fig:eigenmeans}). Because of this behavior, in the tables and figures below, we only present the data from a low mean, or dark, data set, $\cD_{.13}$, the data from a moderate meaned data set $\cD_{norm}$ and the data from a high meaned, or bright, data set , $\cD_{5, .8}$.

Visually, there are two things to note. First, for a given autoencoder with a large enough latent dimension, the reconstructed image appears be a generalized character, not a one from the MNIST data set. For lower latent dimensions, the reconstructed images appear uniformly to be a character from the MNIST data set. For very low latent dimensions, the 10 distributions are reconstructed as elements of the MNIST data set, but not uniformly. Visual similarity of the reconstruction images for each autoencoder implies that one may think of the image of all these data manifolds as lying in a subspace of $\cR$, $M_{dist, d}$, where $d = \dim(\cL)$. That is, for $f_{enc, d}$ and $f_{dec, d}$ the autoencoder with $d$ dimensional latent space, we write \bas f_{dec, d} \circ f_{enc, d}: M_{\cD_*} \rightarrow M_{dist, d} \eas for $M_{\cD_*}$ corresponding to the Bernoulli, Beta or truncated normal distribution manifolds. It is striking to note that the manifold $M_{dist,d}$ is a function of $d$, i.e. different for each architecture. Furthermore, it appears to be different for each seed.

We posit that some of this observed behavior is due to the fact that at very low latent dimensions, there are not enough degrees of freedom in the latent space for the autoencoder to have learned enough features to be able to do anything but noisily assign out of distribution objects similarly. As the degrees of freedom increase, the autoencoders seem to first be able to identify the out of distribution elements as similar. However, these systems do not have the potential to reconstruct images as anything other than those that are in distribution. Eventually, for high enough $d$, the autoencoder has enough degrees of freedom to be able to reconstruct the out of distribution image as a different out of distribution image. How it does it, however, is my no means consistent across systems.

\subsection{Eigenvalue behavior}

\begin{figure} 
\centering
\begin{subfigure}[t]{0.2\textwidth}
\includegraphics[width=\textwidth]{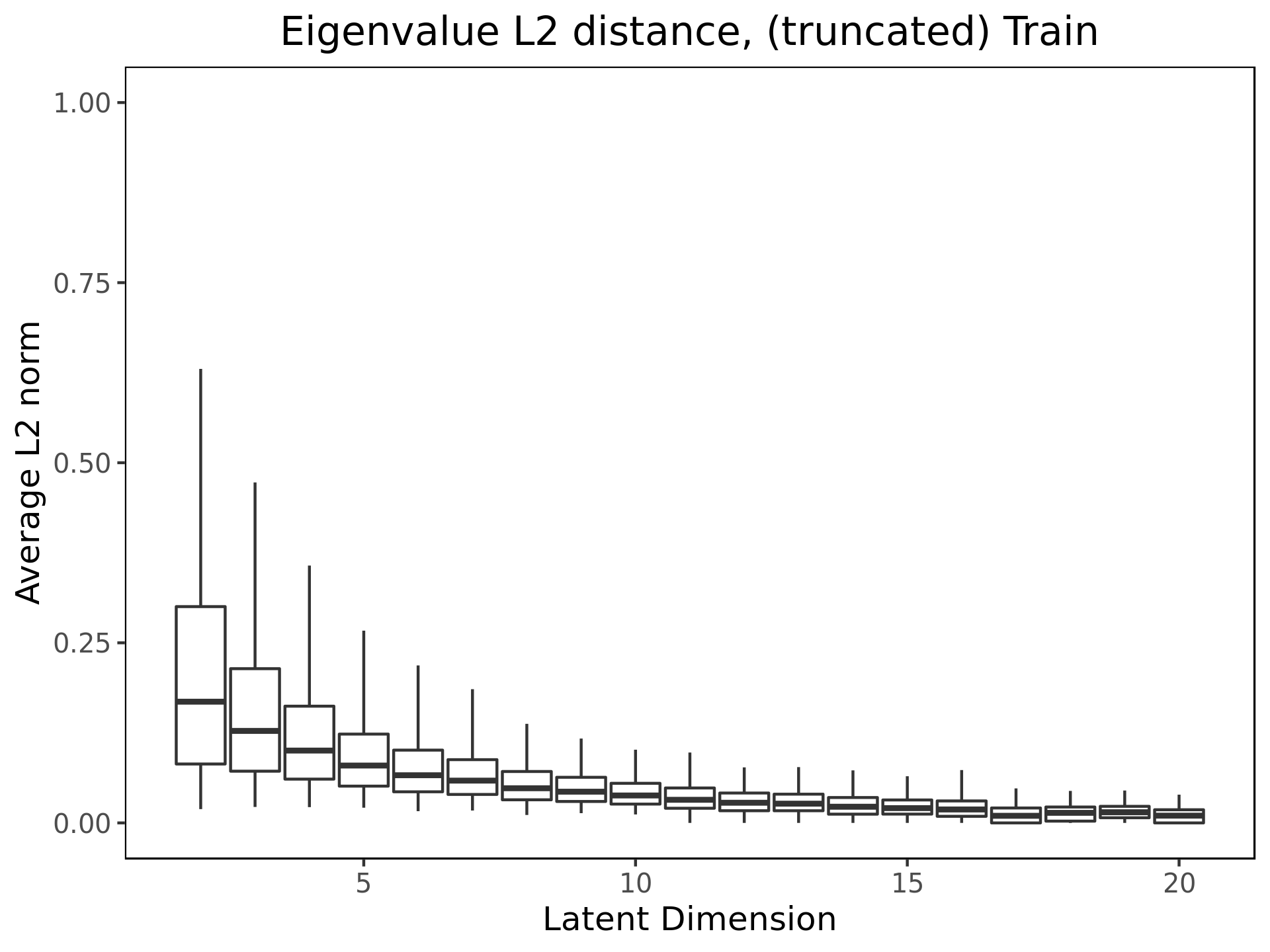}
\end{subfigure}
\begin{subfigure}[t]{0.2\textwidth}
\includegraphics[width=\textwidth]{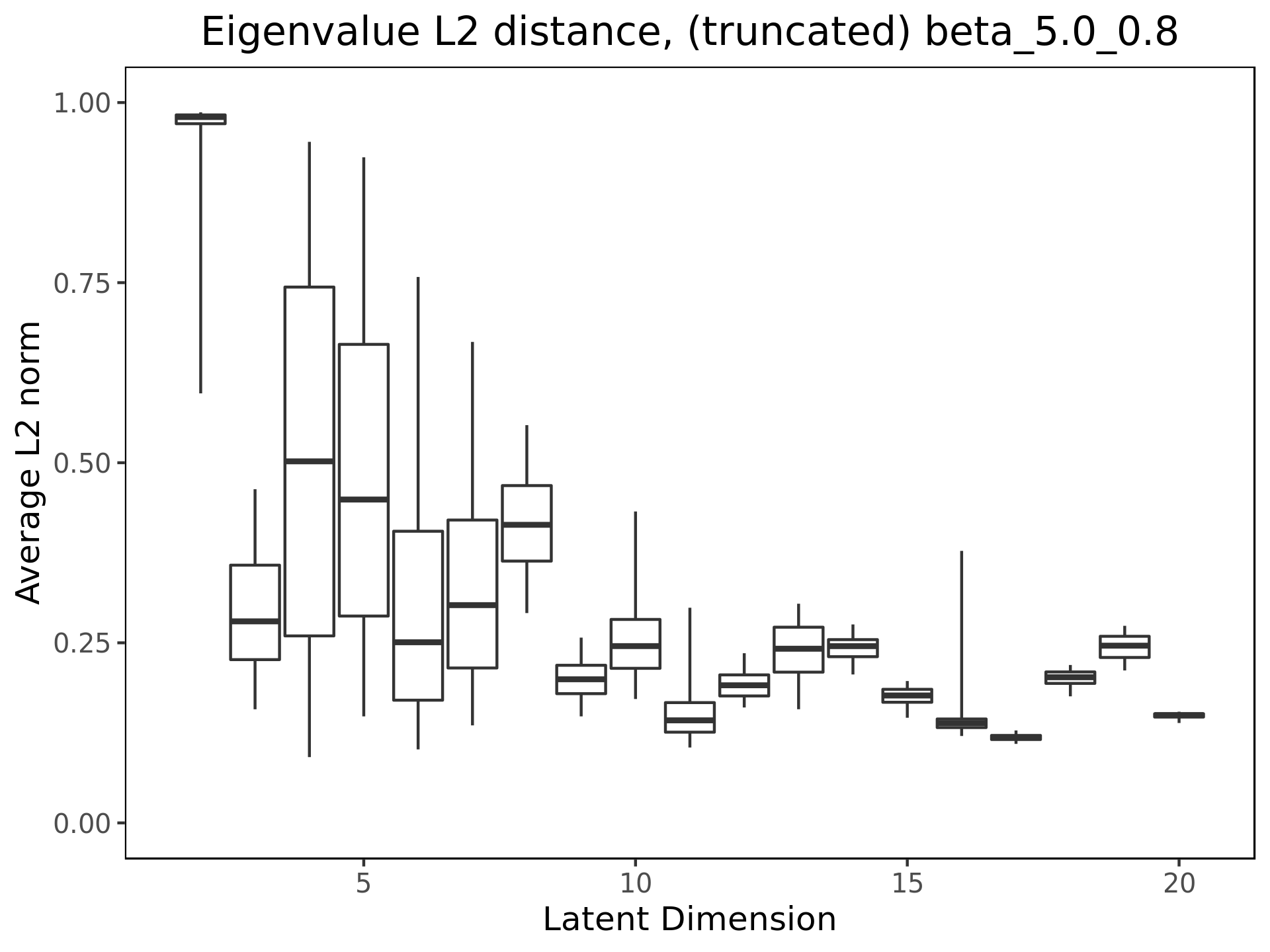}
\end{subfigure}
\par 
\begin{subfigure}[t]{0.2\textwidth}
\includegraphics[width=\textwidth]{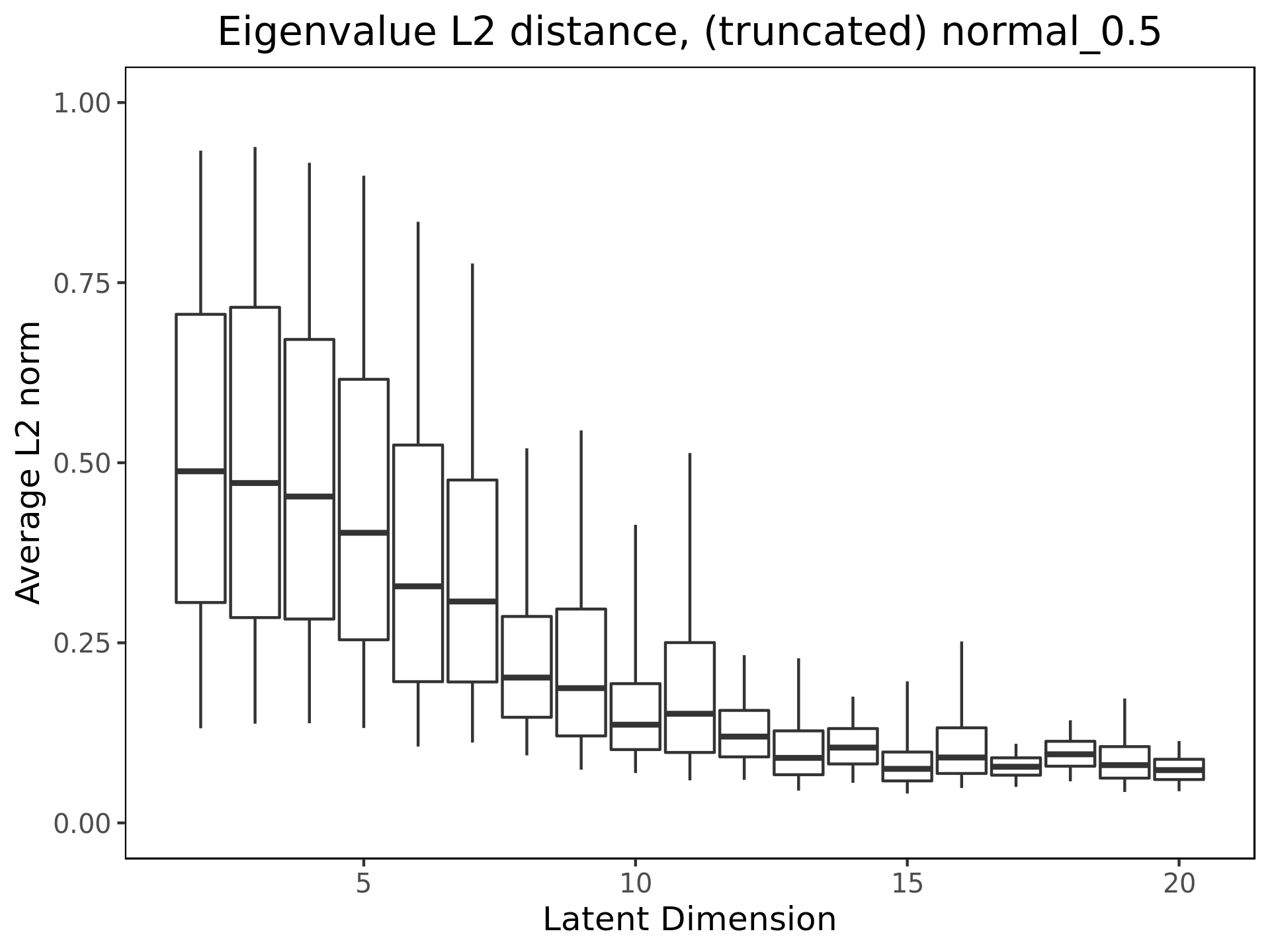}\end{subfigure}
\begin{subfigure}[t]{0.2\textwidth}
\includegraphics[width=\textwidth]{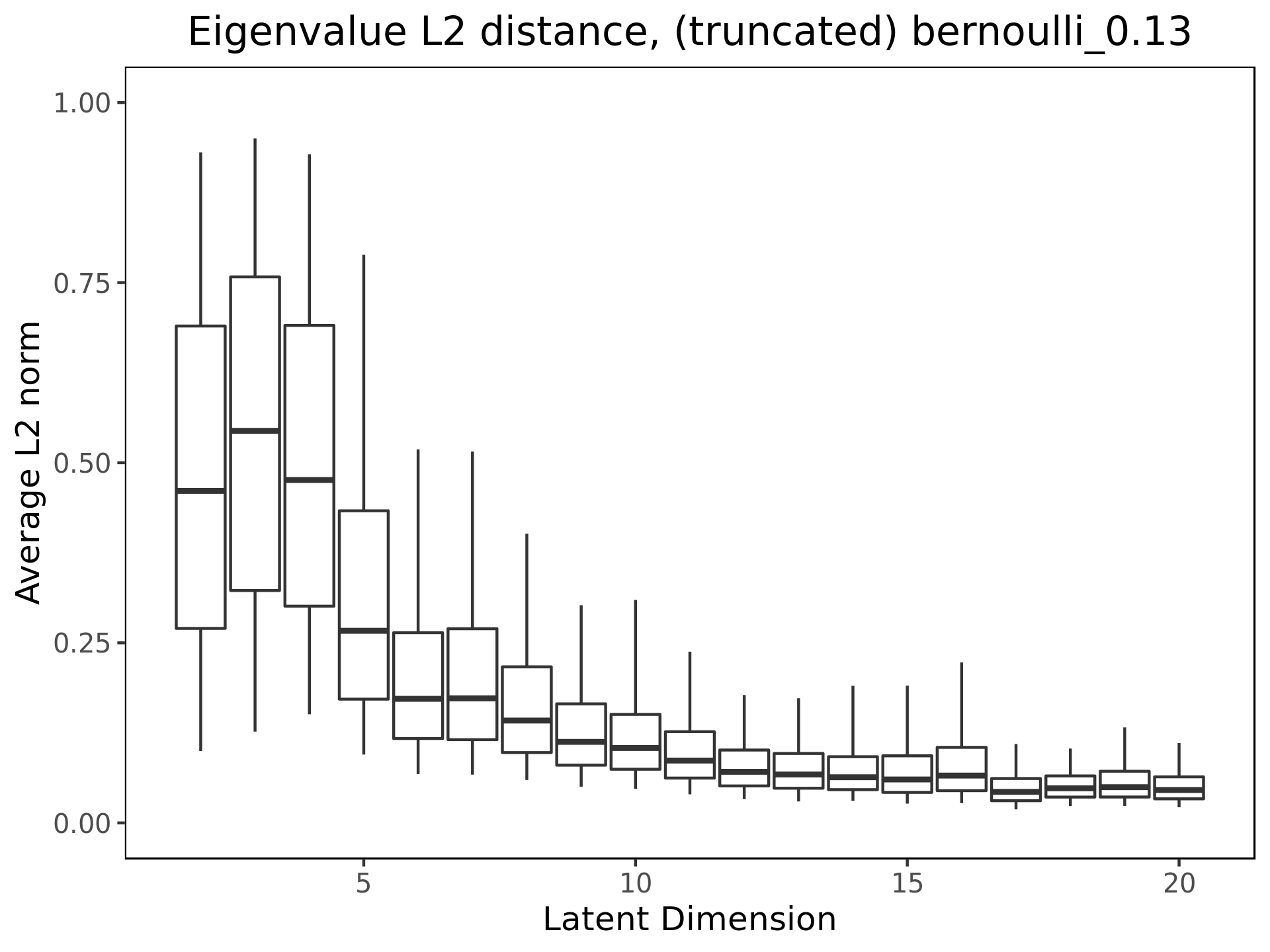} 
\end{subfigure}

\caption{The ratio of the $L_2$ norm of the difference in eigenvalues to the latent dimension for $\cD_{train}$, $\cD_{5, .8}$, $\cD_{normal}$ and $\cD_{.13}$. The reconstruction errors are only small for $\cD_{train}$, therefore, the distance $\|\vec{\lambda}_{J_\cI(x)} - \vec{\lambda}_{J_\cL(z)}\|^2/d$ is larger the out of domain data.} \label{fig:eigennorms}
\end{figure}

Next, we study the behavior of the eigenvalues of $J_\cI(x)$ and $J_\cL(z)$. By Theorem \ref{res:evalssame}, if the reconstruction error is small, we expect the vectors $\vec{\lambda}_{J_\cI}(x) - \vec{\lambda}_{J_\cL}(z)$o be small. When the reconstruction error is large, we expect to see no such pattern. This is borne out in Figure \ref{fig:eigennorms}. 

\begin{figure} 
\centering
\begin{subfigure}[t]{0.2\textwidth}
\includegraphics[width=\textwidth]{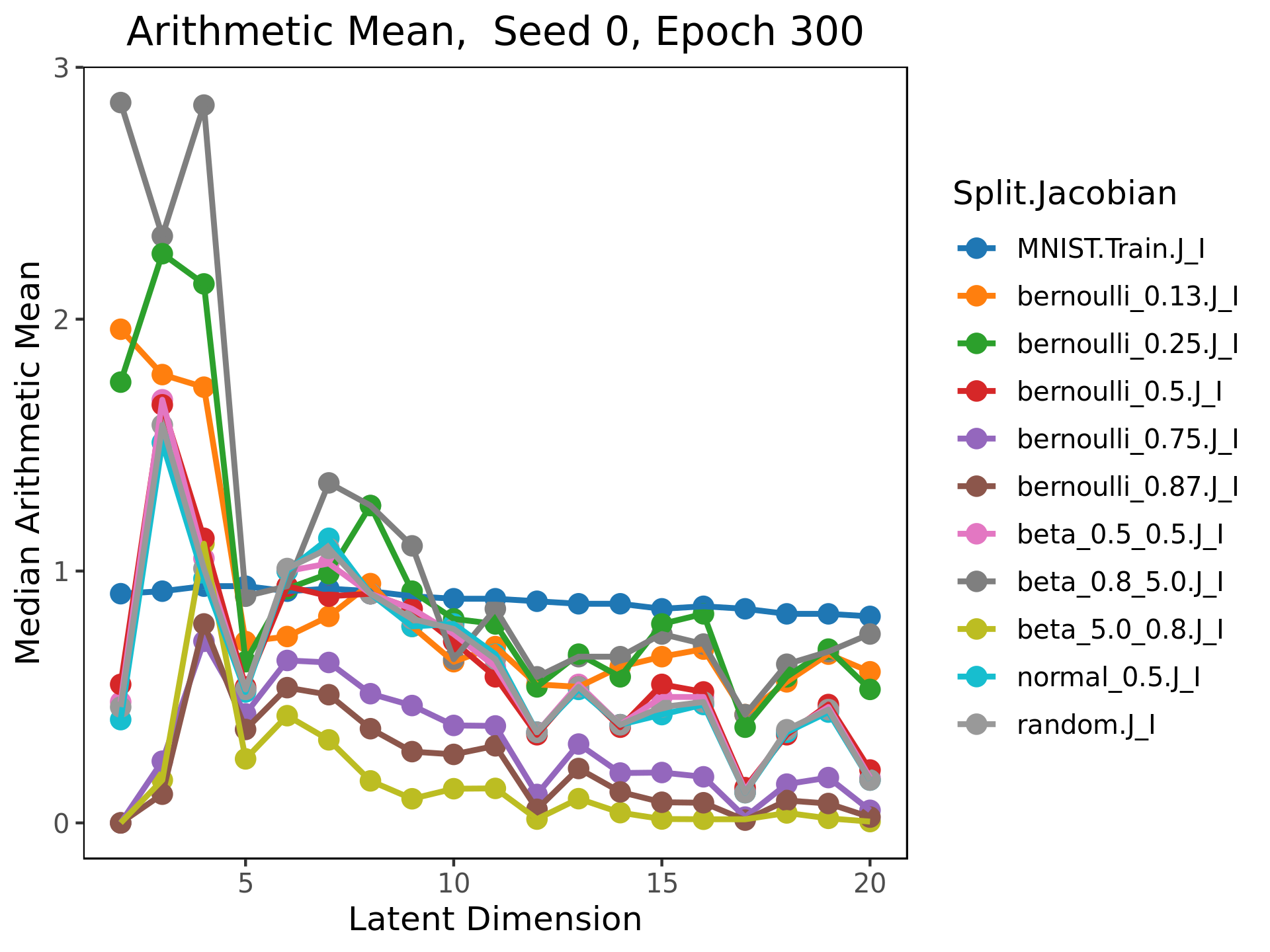}
\end{subfigure}
\begin{subfigure}[t]{0.2\textwidth}
\includegraphics[width=\textwidth]{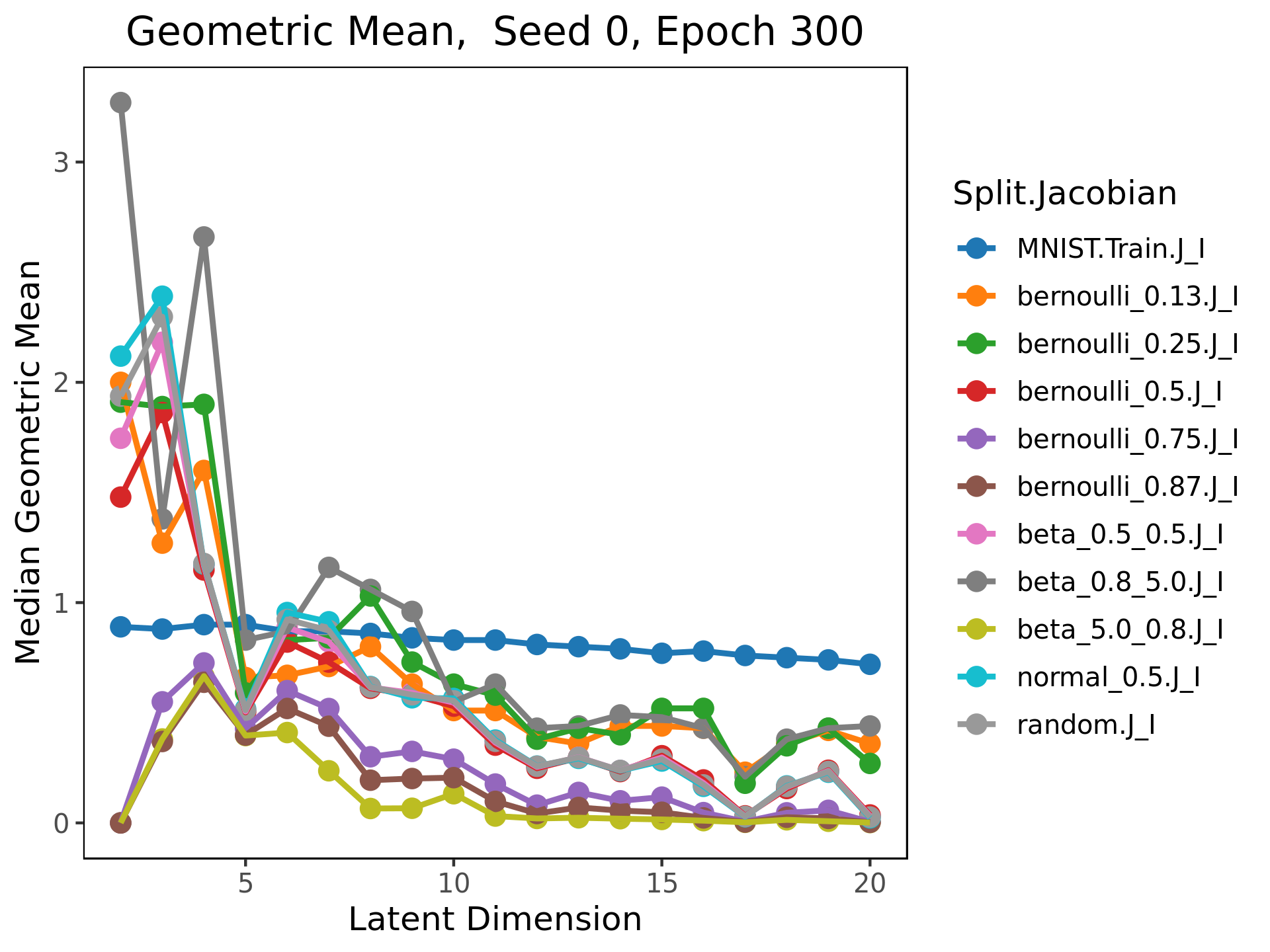}
\end{subfigure}
\par 
\begin{subfigure}[t]{0.2\textwidth}
\includegraphics[width=\textwidth]{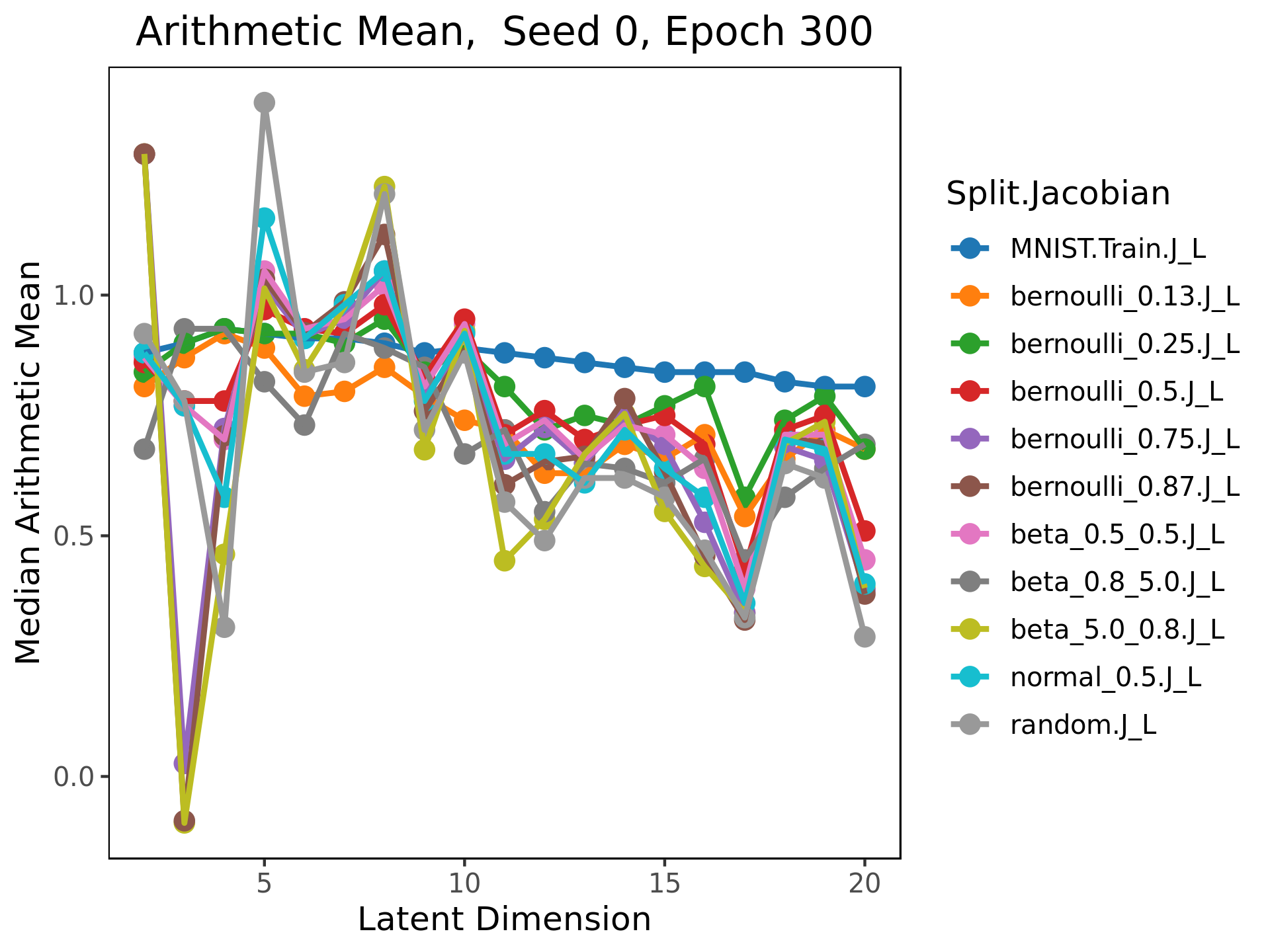}\end{subfigure}
\begin{subfigure}[t]{0.2\textwidth}
\includegraphics[width=\textwidth]{graphs_from_abyss/graphs_seed-0_epochs-300_random_collective_L/exp-default/arith_median.png} 
\end{subfigure} 
\caption{The arithemetic and geometric means for $\cD_{train}$ and the 10 out of distribution evaluation sets. Eventually, the mean eigenvalues for the out of distribution data sets drop below those of $\cD_{train}$. The mean $\vec{\lambda}_{J_\cI(x)}$ cluster for the out of distribution data sets in a way that the $\vec{\lambda}_{J_\cL(x)}$ does not. }\label{fig:eigenmeans}
\end{figure}

\begin{figure} 
\centering
\begin{subfigure}[t]{0.2\textwidth}
\includegraphics[width=\textwidth]{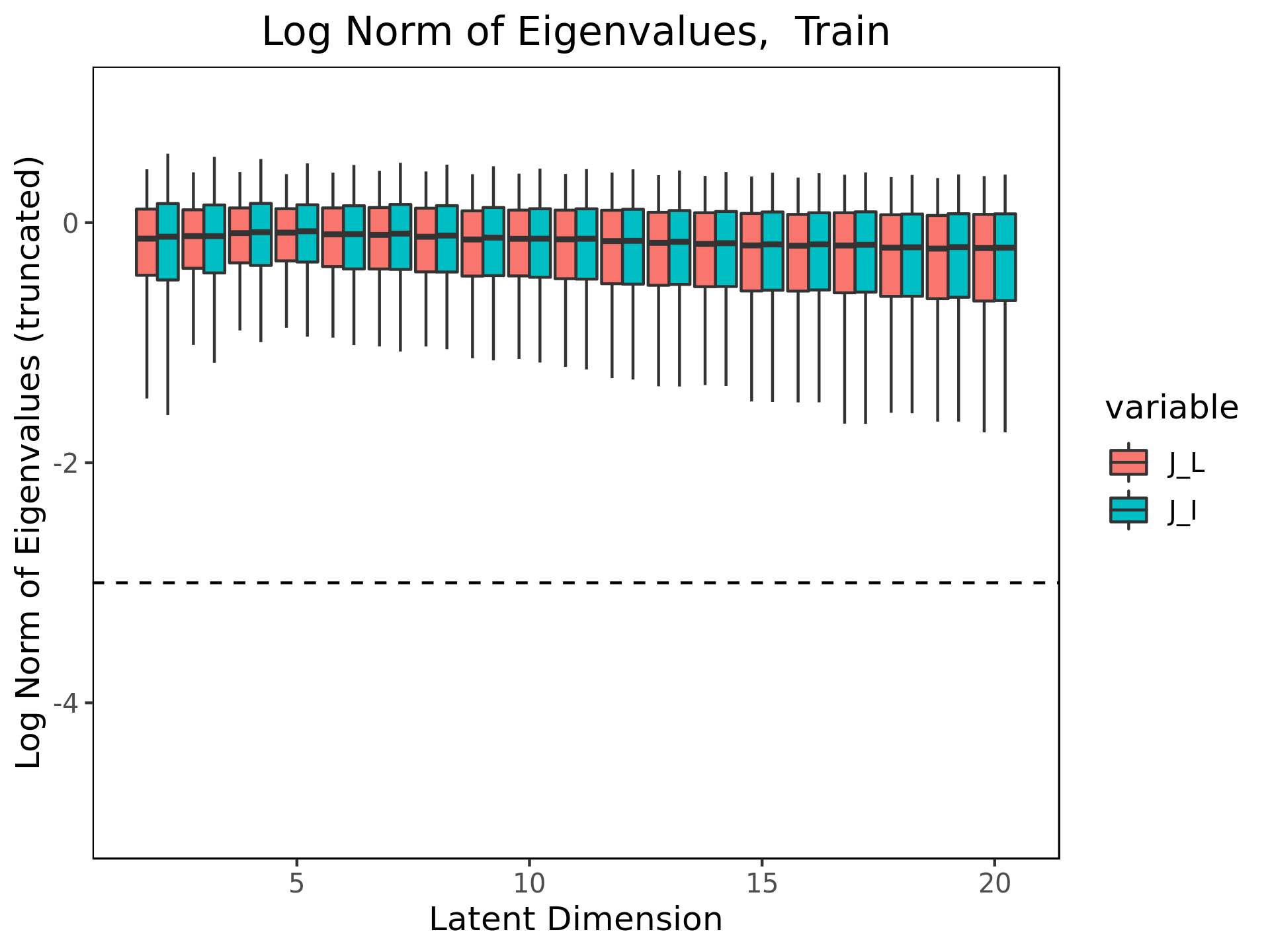}
\end{subfigure}
\begin{subfigure}[t]{0.2\textwidth}
\includegraphics[width=\textwidth]{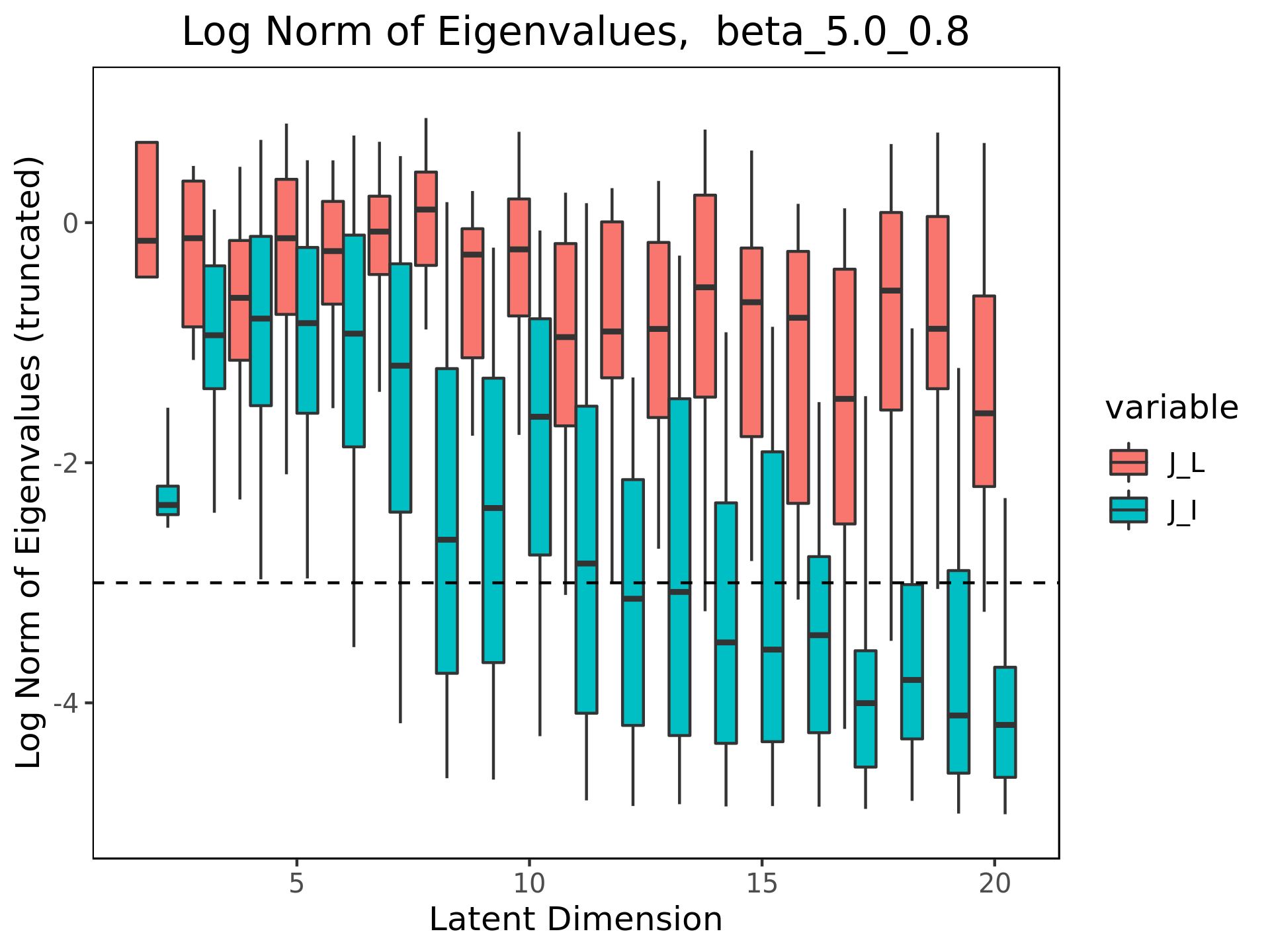}
\end{subfigure}
\par 
\begin{subfigure}[t]{0.2\textwidth}
\includegraphics[width=\textwidth]{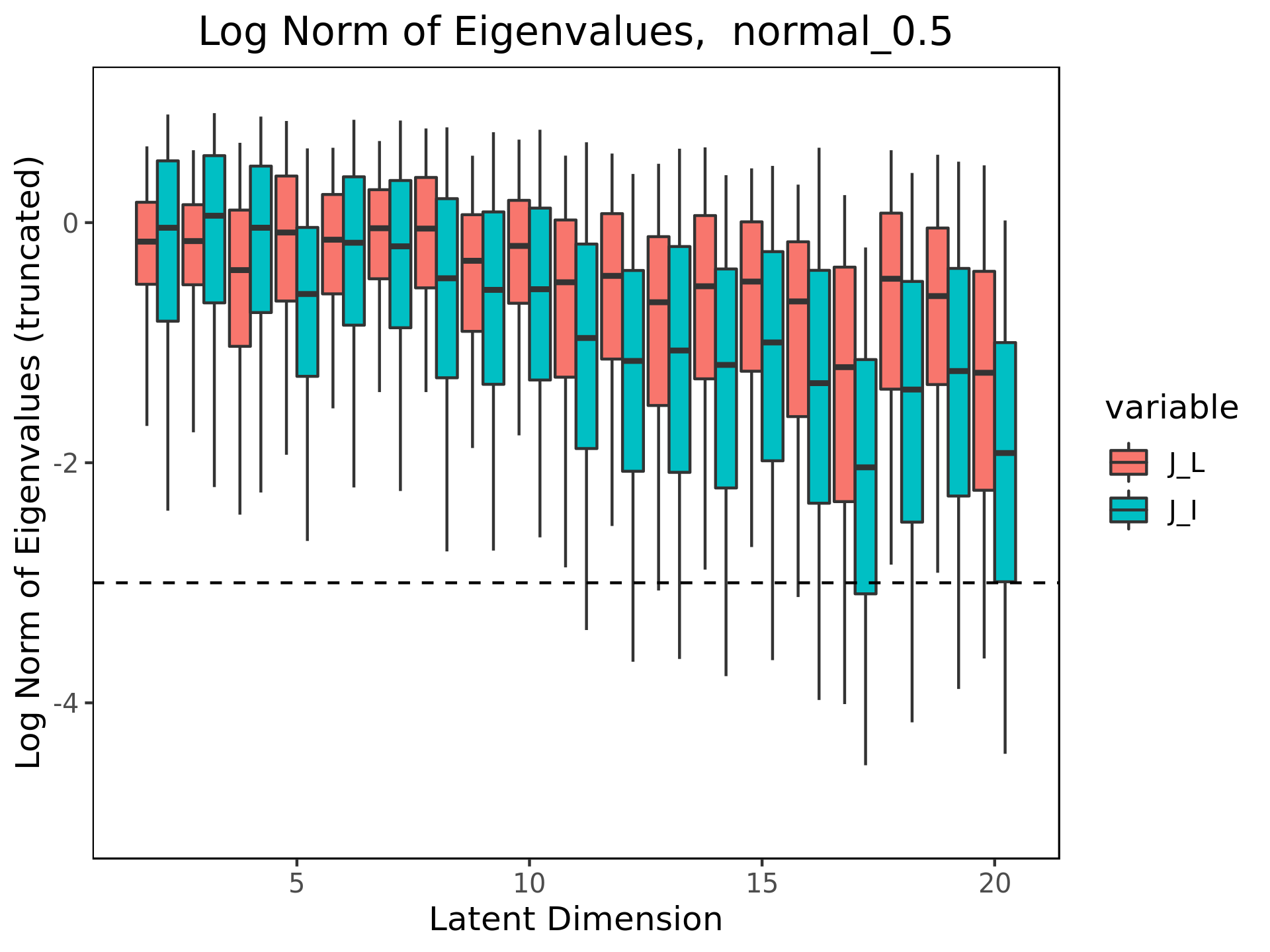}\end{subfigure}
\begin{subfigure}[t]{0.2\textwidth}
\includegraphics[width=\textwidth]{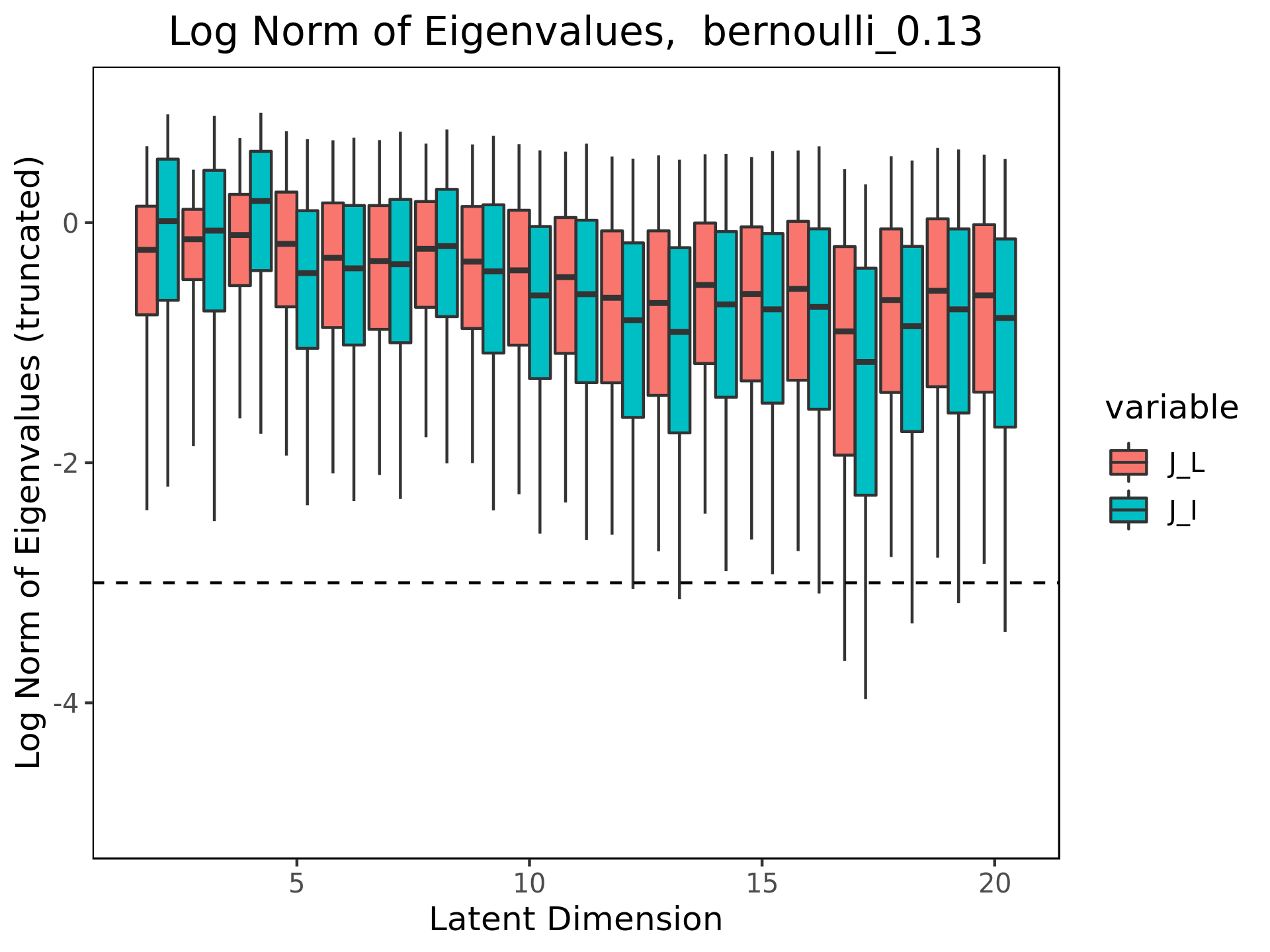} 
\end{subfigure}

\caption{The box plots for $log(|\vec{\lambda}_{J_\cI(x)})_i|$ (blue) and $\log(|\vec{\lambda}_{J_\cI(x)})_i|$ (red). The medians of the distributions any given latent dimension varies more than for the in distribution data. } \label{fig:lognormbox}
\end{figure}

For the in distribution data, the arithmetic and geometric means of the eigenvalues are near $1$, with a slow decline as the latet dimension increases (see Figure \ref{fig:eigenmeans}). The similarity of scale between the arithmetic and geometric means indicates that eigenvalues do not contain extreme outliers at any latent dimension. As discussed in Section \ref{sec:indisttheory}, this indicates that the autoencoder is learnining a function that is close to a projection onto the data manifold. At high latent dimenion, the autoencoder is not learning to ignore certain dimensions of the data (as evidenced by the failure of $0$ eigenvalues to appear in significant quantities, see  \ref{fig:lognormbox}), but the decrease in mean eigenvalue indicates that the autoencoder has learned to contract certain dimensions. The fact that, at higher latent dimension, the autoencoder retains some information about all dimensions of the latent space (rather than collapsing them altogether) is consistent with the fact that at higher latent dimension, the out of domain data is reconstructed as generalized characters. In other words, this is exactly the behavior one would expect if the autoencoder is learning (and therefore reconstructing to) more than just the MNIST digits.

For the out of distribution data the arithmetic and geometric means for both $J_\cI$ and $J_\cL$ are far less constant than those of the in distribution data. This architecture dependent variation of the eigenvalues of the autoencoder is evidence of the geometric instability of the autoencoder on out of distribution data. However, for large enough latent dimension, there are three distinct clusters of eigenvalues of $J_{\cI}$: the three darkest distributions(those with the lowest mean pixel value) have the highest average eigenvalue, followed by the distributions with a mean of .5, and the three with the highest mean pixel value have the lowest mean eigenvalue.

Inspite of the apparent learning, most of the eigenvalue information points to a low level of geometric stability on out of domain data. At low latent dimension, when the reconstructed images appear to lie in $M_{\cD_{train}}$ for all out of distribution data sets, the mean eigenvalues of $J_{\cL}$ for many out of distribution data sets are not close to the mean eigenvalue for the training data. By Theorem \ref{res:JLbound}, the squared distance between the vectors of eigenvalues $\|\vec{\lambda}_{J_\cL}(z) - \vec{\lambda}_{J_\cL}(z')\|^2$ is a lower bound for the distance squared between the latent representation $z, z' \in \cL$. A large difference in means (as observed in Figure \ref{fig:eigennorms}) implies that the latent representations of the points in certain data sets are, on average, far away from the latent representations of the points in the training data, even though the reconstructed images are similar. 


\begin{figure} 
\begin{subfigure}[t]{0.2\textwidth}
\includegraphics[width=\textwidth]{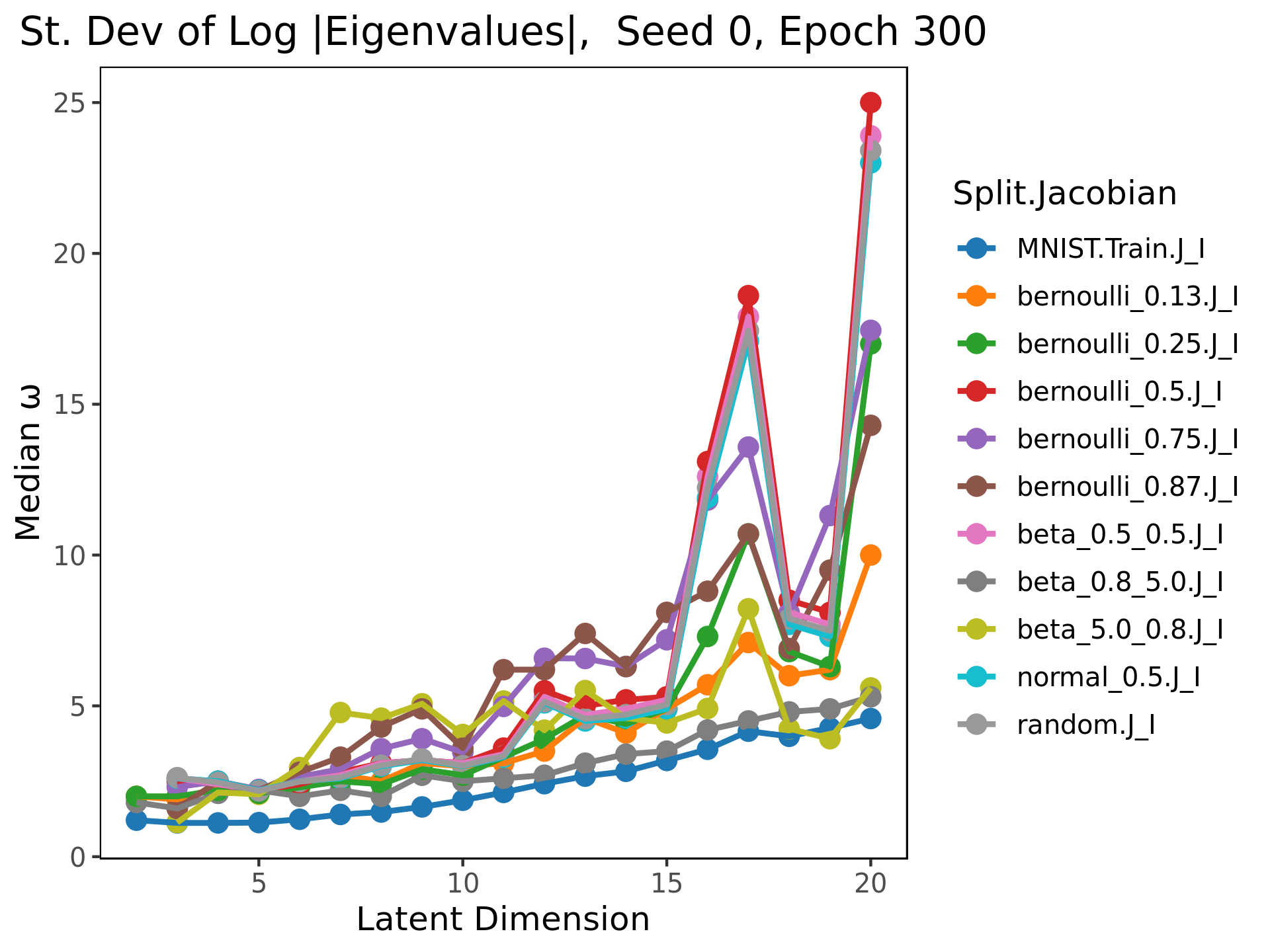}
\end{subfigure}
\begin{subfigure}[t]{0.2\textwidth}
\includegraphics[width=\textwidth]{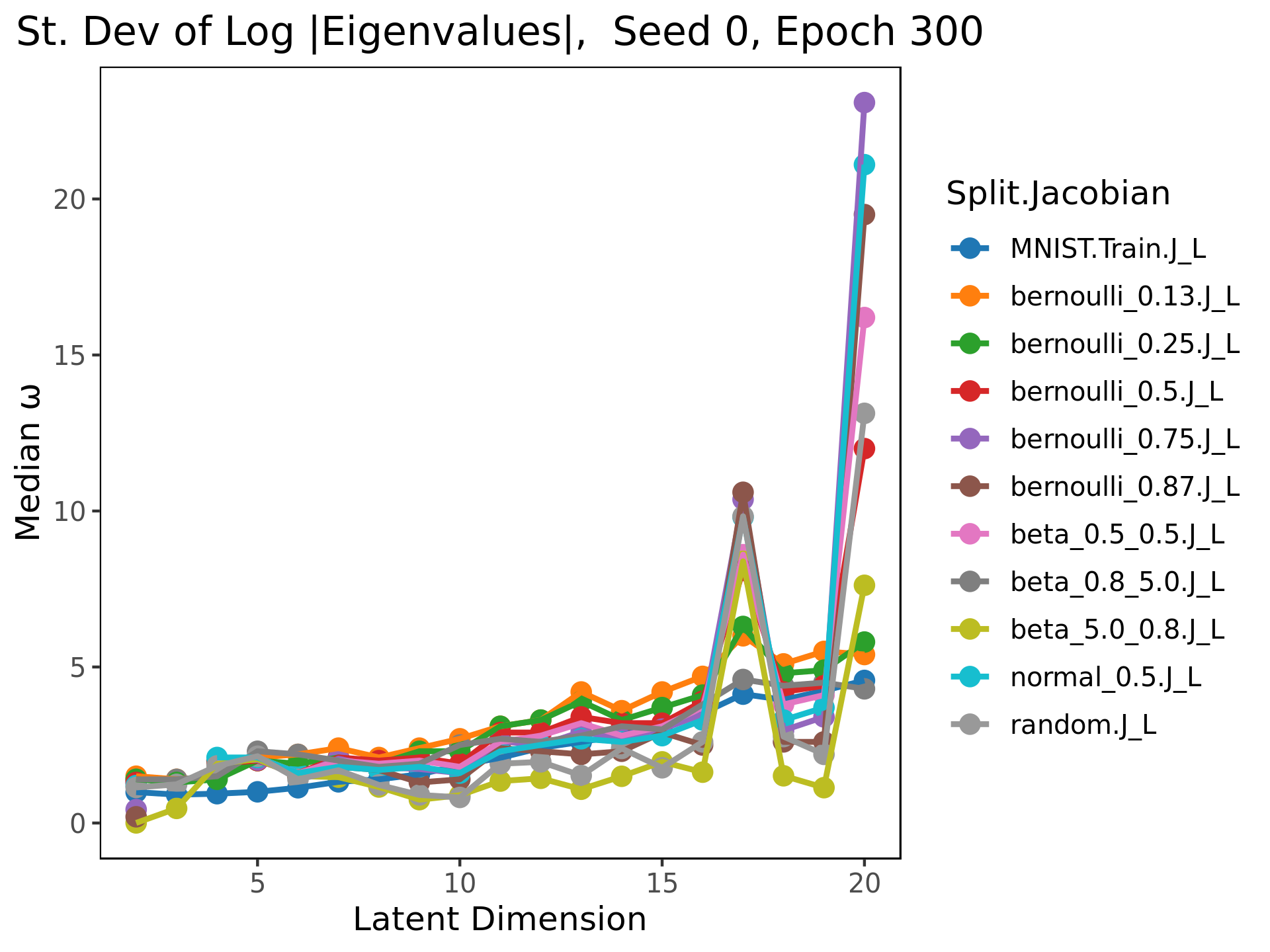}
\end{subfigure}
\caption{The standard deviations of the $\log |\lambda_i|$ distribution for every out of domain data set is higher than the standard deviations for the training data in $J_\cI(x)$, showing the geometric instability of the autoencoder on out of doman data.}\label{fig:stdev}
\end{figure}

Figure \ref{fig:lognormbox} shows the distribution of eigenvalues in the training data, representative bright, moderate and dark data sets. Note that for each of the out of distribution data set, the variance of the distribution of logarithms of the norms of the eigenvalues in any given latent dimension, as well as the medians of the same across latent dimension is much greater than the same variance for the in distribution data (see Figure \ref{fig:stdev}). This again points to the geometric instability of the autoencoder far from the training manifold.

\begin{figure} 
\begin{subfigure}[t]{0.2\textwidth}
\includegraphics[width=\textwidth]{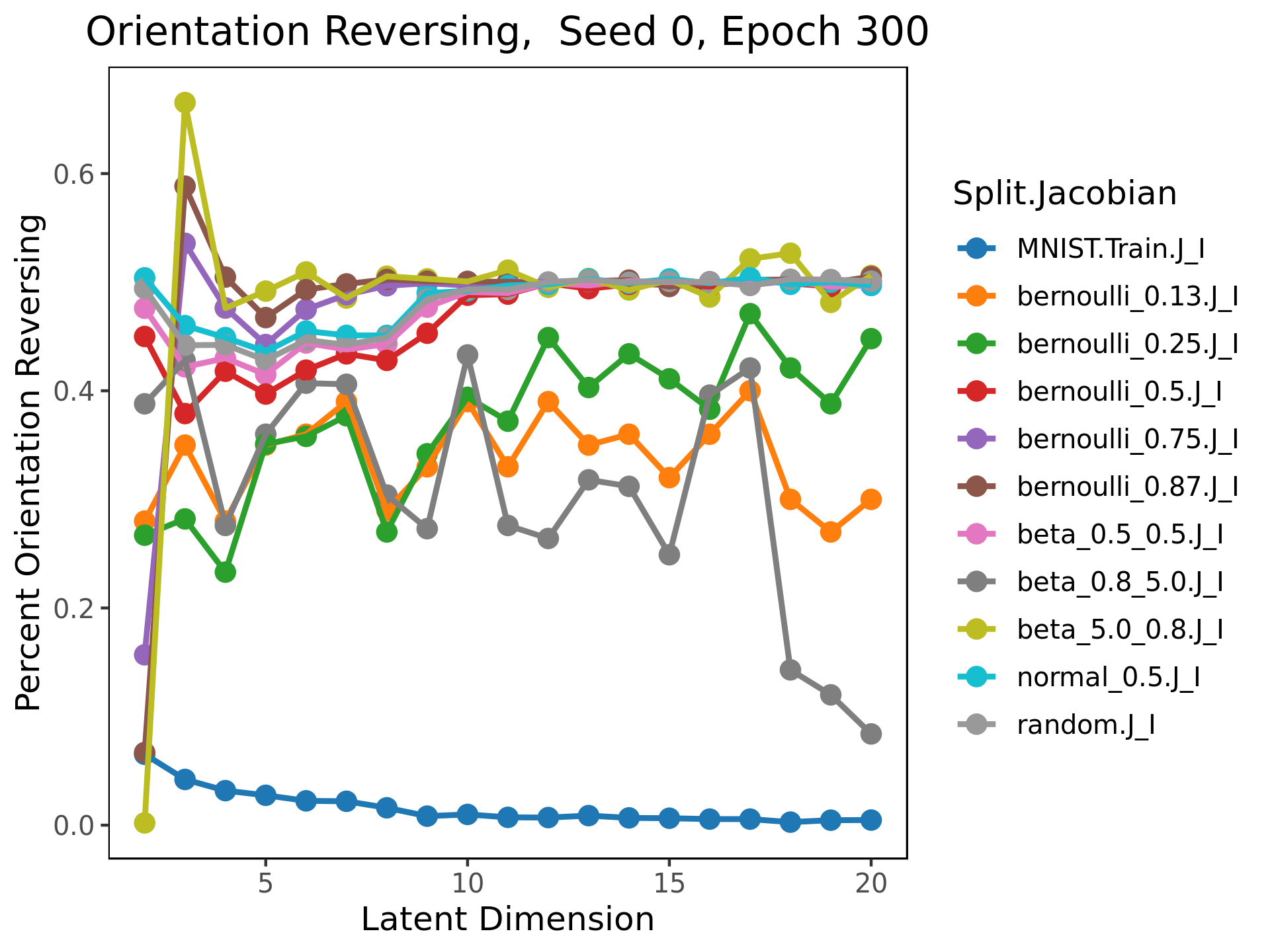}
\end{subfigure}
\begin{subfigure}[t]{0.2\textwidth}
\includegraphics[width=\textwidth]{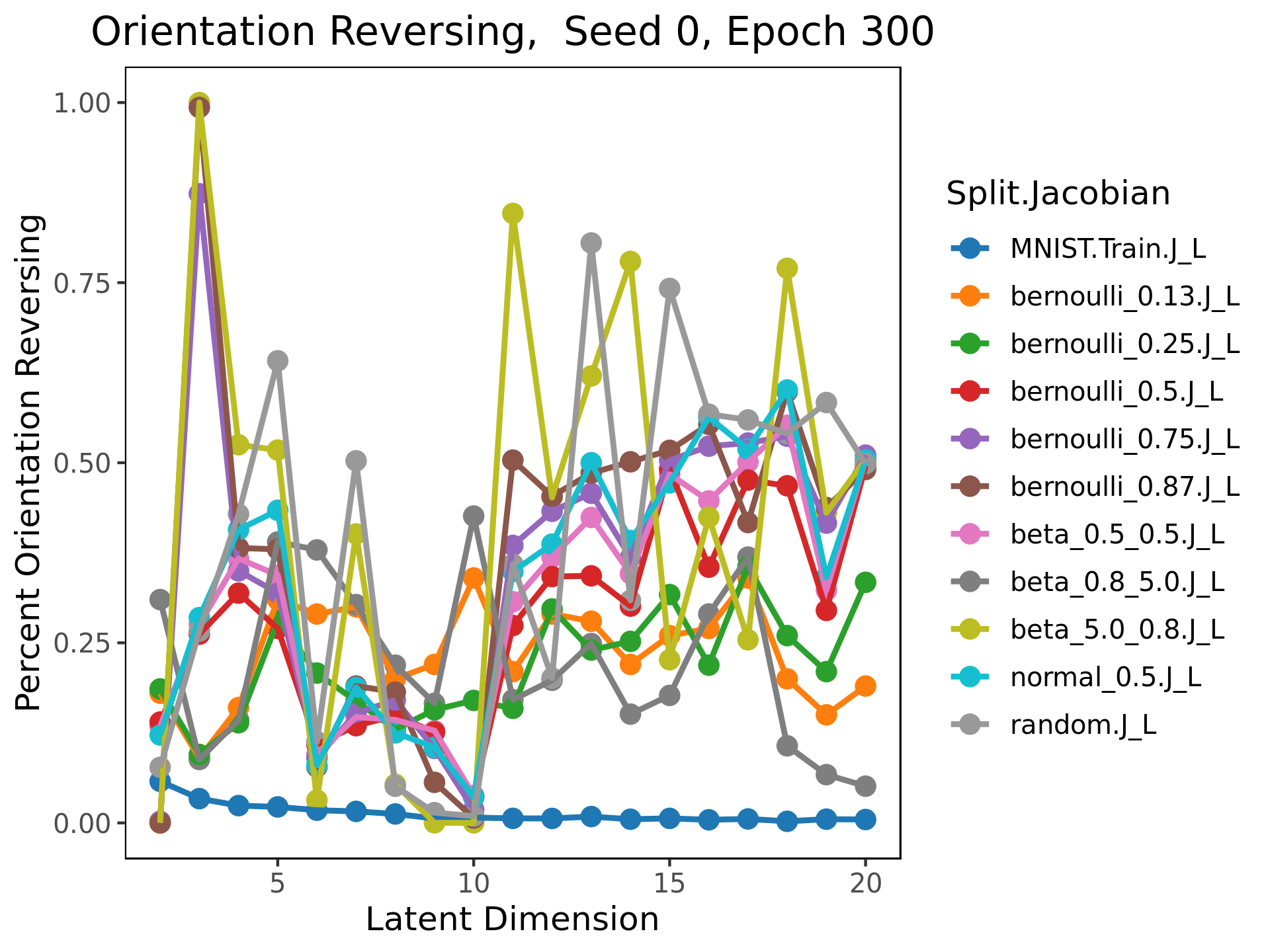}
\end{subfigure}
\caption{The proportion of points for which the product of the top $d$ eigenvalues of $J_\cI(x)$ (left) or $J_\cL(z)$ (right) is negative, indicating that orientation of coordinate system has been flipped. There are very few of these points for $\cD_{train}$, while for the out of distribution data, the proportions are near half for $J_\cI(x)$.}\label{fig:orientation}
\end{figure}

\begin{figure} 
\centering
\begin{subfigure}[t]{0.2\textwidth}
\includegraphics[width=\textwidth]{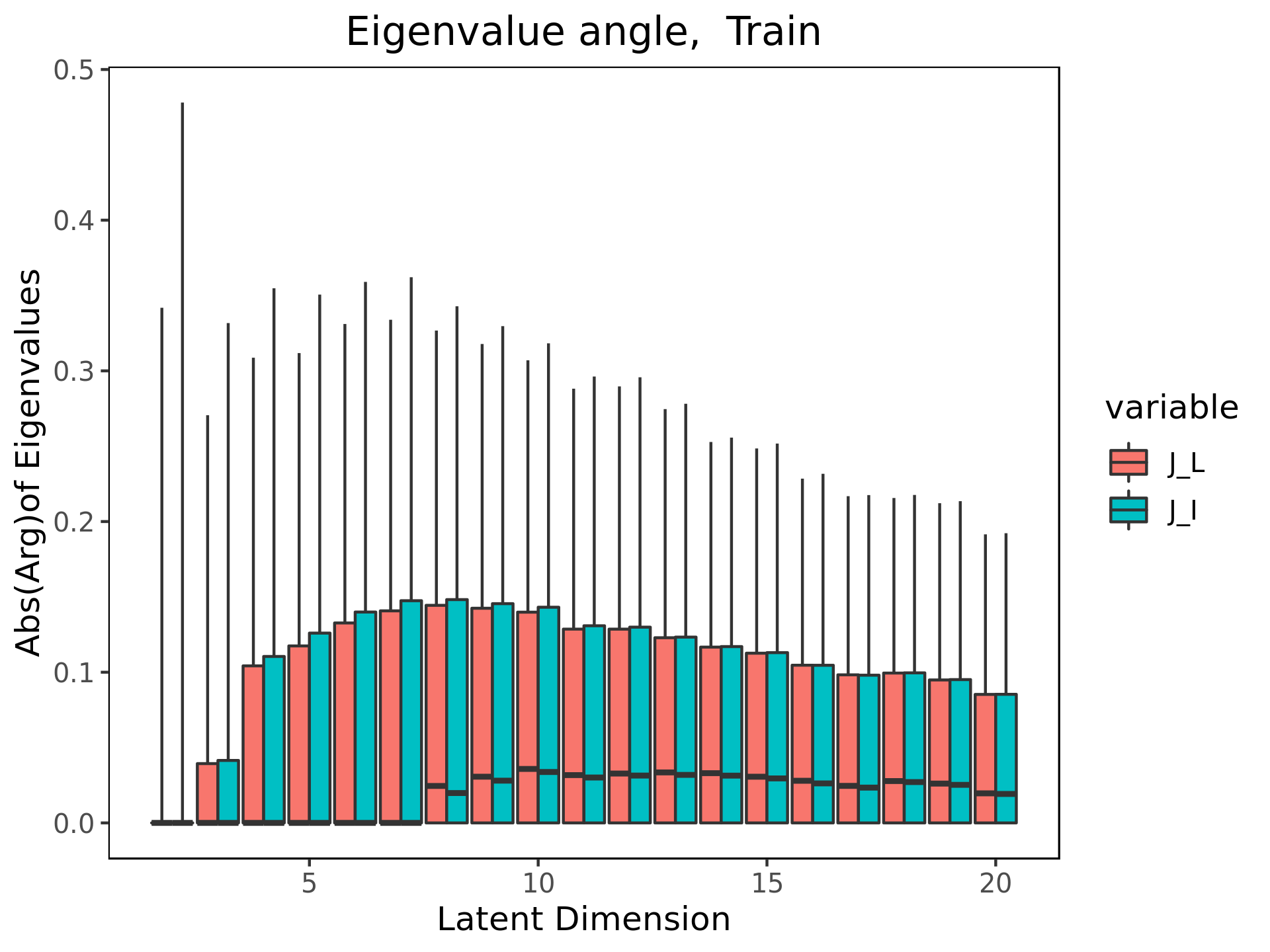}
\end{subfigure}
\begin{subfigure}[t]{0.2\textwidth}
\includegraphics[width=\textwidth]{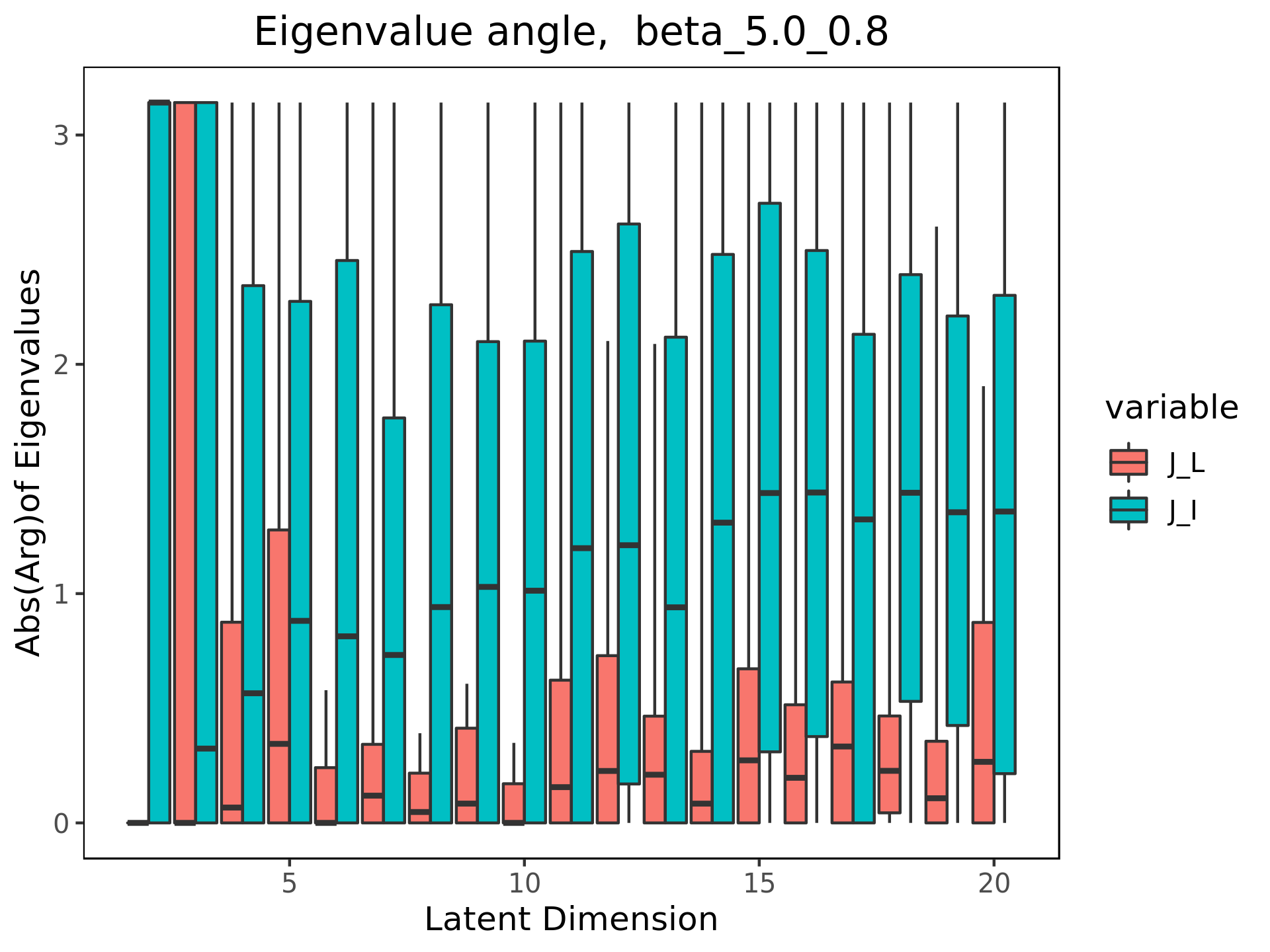}
\end{subfigure}
\par 
\begin{subfigure}[t]{0.2\textwidth}
\includegraphics[width=\textwidth]{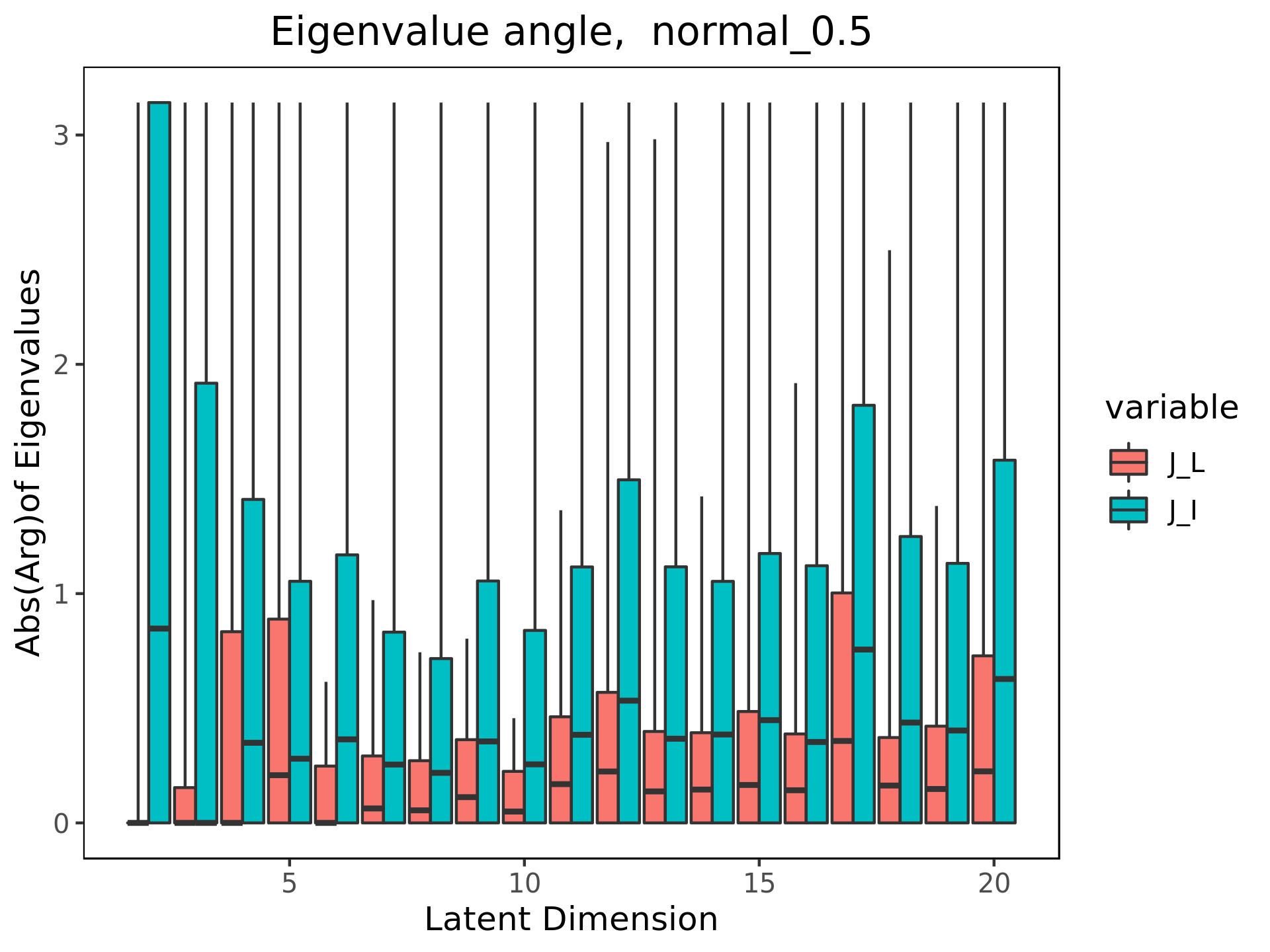}\end{subfigure}
\begin{subfigure}[t]{0.2\textwidth}
\includegraphics[width=\textwidth]{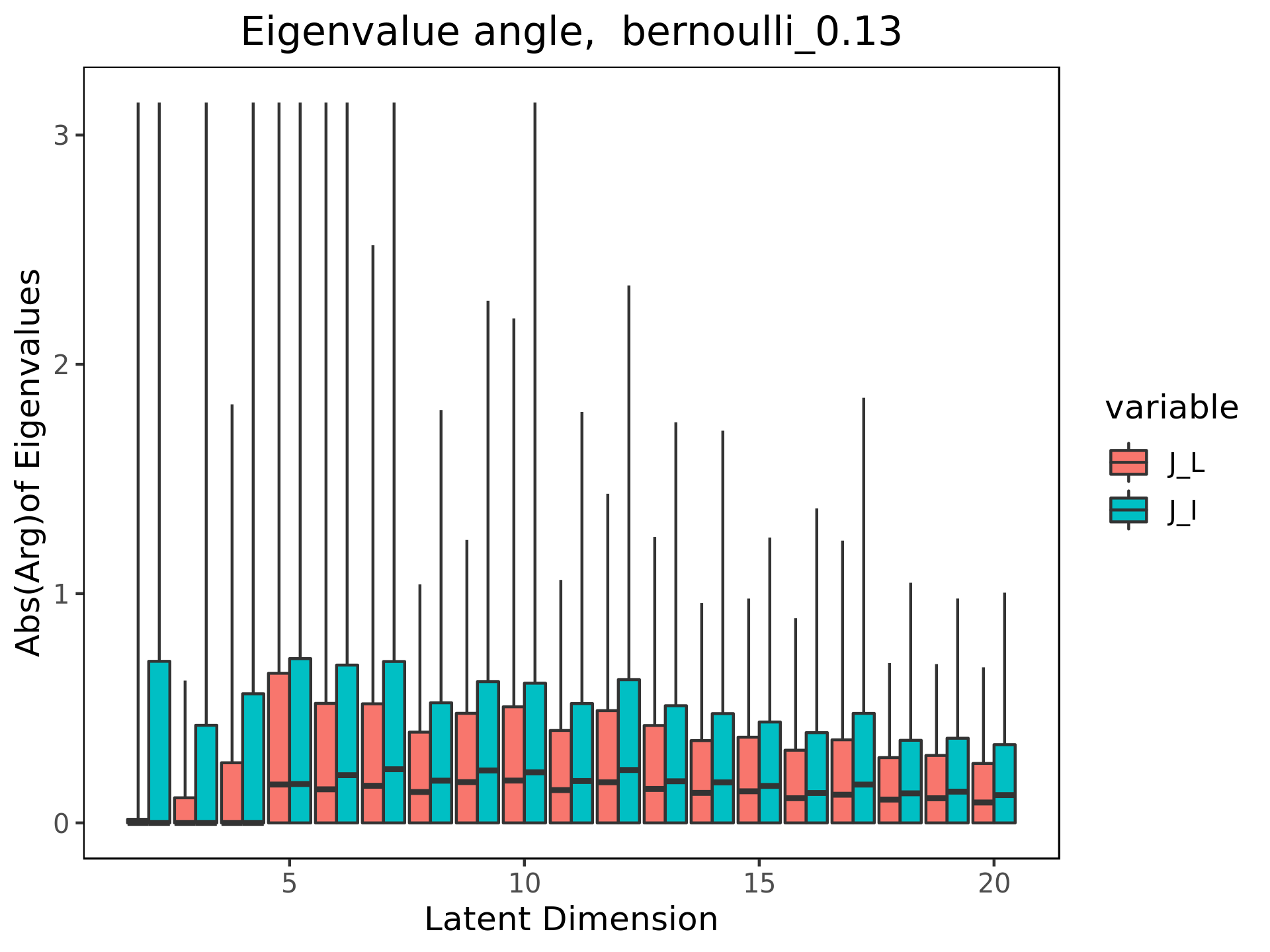} 
\end{subfigure}

\caption{The box plots for $|Arg(\vec{\lambda}_{J_\cI(x)})|)$ (blue) and $|Arg(\vec{\lambda}_{J_\cI(x)})|)$ (red). Note these arguments cover a significantly wider range for the out of distribution data sets than the training data.} \label{fig:anglebox}
\end{figure}

Finally, we note that, by Theorem \ref{res:factor} autoencoder does not map points in $\cI$ far from the training manifold onto $M_{\cD_{train}}$ via an orientation preserving, non-rotating map (such as a parallel transport in $\cI$ in the Euclidean metric). Notably, on the training data, the distribution of arguments of the eigenvalues is small (see Figure \ref{fig:anglebox}) as is the proportion of points for which the orientation is reversed (see Figure \ref{fig:orientation}). However, for the out of distribution data, the arguments for $J_\cI(x)$ can be any angle for the bright data sets. The range of possible angles decreases with range, but even for the dark data sets, is consistently larger than for the training data. Similarly, for the all but the darkest data sets, at high enough latent dimension, the matrices $J_{\cI}(x)$ reverse the orientation of $T_x(\cI)$ approximately half the time. Given the general instability across architectures of other geometric properties on out of distribution data, the consistency of this coin flip on the orientation of the tangent space is remarkable.

\section{Comparison with related work}
The machine learning community is not a stranger to the differential geometry literature. There is a large body of work on learning metric functions on the latent spaces for (variational) autoencoders and to determine the reconstruction loss \cite{arvanitidis2017latent, arvanitidis2020geometrically, eklund2019expected, hauberg2018only, tosi2014metrics}. Others propose building the reconstruction error out of the Mahalanobis distance function \cite{denouden2018improving} to better detect out of distribution inputs. These, and other efforts, aim to use differential geometry tools to improve the performance of neural networks. Our work differs from these in that we are not interested in performance enhancing methodologies, but in understanding the geometry feature map so that we may explain how the feature map perceives and interacts with out of domain data.

We note that Contraction Autoencoders (CAEs) use the Frobenius norm of the Jacobian of the encoder map to regularize the loss function \cite{rifai2011contractive}. CAEs and their higher order counterparts \cite{rifai2011higher} are better at capturing smoother, more accurate and lower dimensional representations of the data manifold. As mentioned above, we explicitly avoid regularization techniques in the autoencoders trained for this paper in order to better understand the geometry of the learned functions. Furthermore, the Jacobian matrices used in these cases were of the encoder maps only, while our work focuses on the Jacobian matrix for the two networks combined.

\section{Conclusion}

In this paper, we study the learned geometry of a family of trained autoencoders, on input points far away from the data set on which the system was trained. Instead of studying the performance outputs of the autoencoders (which we expect to be poor) we look at the geometric features of the autoencoder, viewed as a function from input to reconstruction space. In doing so, we make the following observations. From the point of view of the reconstructed images, at low latent dimension the data sets of randomly distributed pixel values are all mapped to images that appear to come from the training data set. However, which class of training image is reconstructed is not consistent neither within random data set, nor across random data sets, nor across seeds. At higher latent dimensions, the reconstructed images across all data sets in a given architecture and seed appear to be a \emph{generalized characters}. However, this character is not consistent across architecture or seed. 

From a geometric perspective, we show that while, for lower latent dimension, the reconstructed images appear to be the same as those in the training data, the latent representations of the training and out of distribution points are quite different. Furthermore, we find evidence that that geometries of the learned functions (as measured by the eigenvalues of $J_\cI(x)$ cluster according to the mean value of the pixel distribution. Finally, and possibly most surprisingly, we observe that for higher latent dimension, the trained autoencoder is not aware of the orientation of the bases of the input space.

We contend that studying the geometry of the trained autoencoder will give insight into \emph{how} the network is processing both in and out of distribution data, and will play a key role in explaining the phenomenon of out of distribution reconstruction.

\appendix

\section*{Computing the Jacobians \label{app:Jacobean}}

Explicitly, for the autoencoder structure used in this paper we may write $f_{enc}$ as a composition of the weight matrices and activation functions at each layer. That is, \bmls f_{enc} = f_{enc, 4} \circ f_{enc, 3} \circ f_{enc, 2} \circ f_{enc, 1} \\ f_{dec} = f_{dec, 4} \circ f_{dec, 3} \circ f_{dec, 2} \circ f_{dec, 1} \emls where
\bas f_{enc, i}(x) &= 
\begin{cases}
ReLU_{128}(A_{128 \times 784} x + b_{128}) & i = 1 \\ 
ReLU_{64}(A_{64 \times 128} x + b_{64} ) & i = 2 \\
ReLU_{32}(A_{32 \times 64} x + b_{32} ) & i = 3 \\
A_{d \times 32} x + b_d  & i = 4 \;,
\end{cases}
\quad
\\
\quad f_{dec, i}(x) & = 
\begin{cases}ReLU_{32}(A_{32 \times d} x + b_{32}) & i = 1 \\
ReLU_{64}(A_{64 \times 32} x + b_{64} ) & i = 2 \\
ReLU_{128}(A_{128 \times 64} x + b_{128} ) & i = 3 \\
\tanh_{784}(A_{784 \times 128} x + b_{784})  & i = 4 \;,
\end{cases} \; .
\eas
Here, each $A_{n \times m}$ is a real $n \times m$ matrix, $b_m$ is a $m$ dimensional column vector, and $ReLU_m : \R^m \rightarrow \R^m$ is the vector valued function that is $ReLU$ in each component (similarly for $\tanh_m$). 

The Jacobians for the autoencoders $J_\cI(x)$ and $J_\cL(z)$ are computed locally by repeated application of the chain rule through both the encoder and decoder layers. Note that the derivative of a layer is applied to the output from the last layer:
\begin{equation*} \label{eq:jacobian}
    F = F_n \circ \cdots \circ F_1, \textrm{ then } \D F  = \prod \D F_i |_{F_{i-1}\circ \cdots \circ F_1(x)} \;.
\end{equation*}
Explicitly the derivative of a affine transformation is the associated matrix is 
\begin{equation*} \label{eq:d_affine}
    \frac{\partial (Ax + b)}{\partial x} = A \;.
\end{equation*}
Recall that the ReLU function is, component wise the piece wise linear function \bas ReLU(x) = \begin{cases} 0 & x <0 \\ x & x \geq 0 \end{cases}\;.\eas The derivative of the ReLU function applied to an $N$ dimensional vector is the Heaviside function embedded along the diagonals of an $N \times N$:
\begin{equation*} \label{eq:d_relu}
    \frac{\partial ReLU(x)}{\partial x} = H(x) =
    \begin{cases}
        1, & x \geq 0 \\
        0, & x < 0
    \end{cases} \;.
\end{equation*} and the derivative of the $\tanh$ function applied to an $N$ dimensional vector is the derivative embedded along the diagonals of an $N \times N$ dimensional matrix is 
\begin{equation*} \label{eq:d_tanh}
    \frac{\partial \tanh(x)}{\partial x} = 1 - \tanh^2(x)
\end{equation*}
Note that in the autoencoder, the ReLU and the $\tanh$ are applied component wise to the outputs of the previous layer. Thus the corresponding term in the Jacobian calculation is a matrix with the functions $H(x)$ or $\tanh(x)$ along the diagonals.




As a particular example, we can compute that the derivative of $f_{enc,1}$ is
\[
\D f_{enc,1}(x)=H(A_{128\times784}x+b)\bI_{128}A_{128\times784}
\]
and the other derivatives can be computed similarly, as can the derivatives of compositions (using the chain rule).

\section*{Geometric results \label{app:geometry}}

\subsection*{Proof of Theorem \ref{res:evalssame}}

\begin{proof}
This is a result of the chain rule.

Since $f_{dec}\circ f_{dec}(x) = x$, let $w$ be an eigenvector of $J_\cI(x)$ with eigenvalue $\lambda$. Then \bas \D f_{dec}|_z\circ \D f_{enc}|_x(w) = \lambda w \;, \eas where $z = f_{enc}(x)$ as above. If $v = \D f_{enc}|_x(w)$ then \bas  \D f_{enc}|_x \circ \D f_{dec}|_z (v) =  \D f_{enc}|_x \circ \D f_{dec} \circ  \\ \D f_{enc}|_x (w) =  \D f_{enc}|_x  (\lambda w) = \lambda(v)\;. \eas
\end{proof}

\subsection*{Topological interpretation of an autoencoder}

\ba \xymatrix{\cM_\cD \supset \cU_\cD \ar[r]^{\phi_\cU} \ar@{^{(}->}[d] & \R^{\delta} \ar[dr]^{\pi} & & \R^{\delta} \ar[dl]_{\pi} & \cU_\cD \ar[l]_{\phi_\cU}  \subset \cM_\cD \ar@{^{(}->}[d] \\ \cI  \ar[rr]^{f_{enc}} &  & \cL \ar[rr]^{f_{dec}}  & & \cR  }  \label{disp:traindiagram}\ea

Display \eqref{disp:traindiagram} shows how an autoencoder learns from a topological point of view. A datamanifold ($M_\cD$) lies with a complicated embedding inside the input space $\cI$. The actual data set ($\cD$) lies on and around $M_\cD$. We say that it is distributed noisily around $M_\cD$, however, we make no statements about the properties (such as higher moments or homoskedasticity) of this distribution. The autoencoder is supposed to approximate $M_\cD$ using piecewise linear maps, and reconstruct this piecewise linear approximation in $\cR$. This learning process is shown in the bottom row. If an  autoencoder has correctly learned the manifold structure of $M_\cD$, then the composition on the bottom row, restricted to the manifold is the identity: \bas f_{dec}\circ f_{dec}(M_\cD) = \mathbb{I} \; . \eas Note, this identity should hold everywhere on $M_\cD$, specificaly, we do not expect it to hold on the $\cD$, which are, with high probability, not on the manifold. In other words, for a point $x$ on $M_\cD$, the Jacobian matrix $J_\cI(x)$ should be the identity on the tangent space of $M_\cD$, $T_x(M_\cD) \subset T_x(\cI)$ and the normal space, $T^\perp_x(M_\cD) \subset T_x(\cI)$, should be its kernel.

If the autoencoder learns the manifold $M_\cD$, then it has learned an atlas on $M_\cD$. Any $\delta$ dimensional manifold is defined by a set of open sets $U_i$ such that their union gives the manifold$\cup_i U_i = M_\cD$, and a set of diffeomorphisms $\phi_i : U_i \rightarrow \R^\delta$. Each pair $(U_i, \phi_i)$ is called a chart on $M_\cD$. The set of charts is called an atlas. If an autoencoder learns the structure of $M_\cD$, it has learned a chart on $M_\cD$, call it $\phi(U)$. The open set $U$ is an open set that contains all the points on the manifold, as well as the projection of the data points off the manifold onto it. The image $\phi(U)$ is in $\R^\delta$. If $d \geq \delta$, this is embedded into the latent space (the map $\pi$ in dispaly \ref{disp:traindiagram}). If $d < \delta$ then the map $\pi$ is a projection of $\phi(U)$ onto $\R^d$.

In actuality, the map $\pi$ is never a projection, or an embedding. The autoencoder learns to fit the data as best it can with the $d$ degrees of freedom that is is given. Therefore, when $d <\delta$, the autoencoder minimizes its loss function, not by projecting onto a lower dimensional space as it would do if it had learnt the manifold structure, but by finding a $d$ dimensional submanifold of $M_\cD$ that better fits the data. Similarly, when $d > \delta$, the autoencoder minimizes its loss function by using the extra dimensions to fit the data. In doing so, it has learned a manifold that is too large, and allows for phenomenon such as out of domain reconstructions and reconstruction to \emph{generalized characters}, as seen in this paper.

\subsection*{Proof of Theorem \ref{res:JLbound}}

\begin{proof}
Recall that the Frobenius norm of an $n\times n$ matrix, $A$, can be written
\[
\|A\|_F^2=\sum_{i=1}^{\min{m,n}}\sigma_i(A)^2
\]
where $\sigma_i(A)$ denotes the $i^{\text{th}}$ singular value of $A$ (in decreasing order).  

Weyl's majorization theorem gives, for any  $p>0$ and any $1\leq k\leq n$ that
\begin{equation}\label{eq:Weyl}
\sum_{i=1}^k|\lambda_i|^p\leq\sum_{i=1}^k\sigma_i^p.
\end{equation} The first inequality comes from setting $p= 2$. 

The equality is a result of Taylor's theorem.
\end{proof}

\subsection*{Proof of Theorem \ref{res:factor}}

\begin{proof}
Display \eqref{disp:factors} gives the commutative diagram if $f_{enc}(x)$ factors through a projection onto $M_{\cD_{train}}$: \ba  \xymatrix{\cI  \ar[rr]^{f_{enc}} \ar[dr]_\pi & &\cL \ar[rr]^{f_{dec}} & & \cR \\ & M_{\cD_{train}}  \ar[ur]_{f_{enc}} &    & &   }  \label{disp:factors} \;,\ea where $\pi$ denotes a paralell transport of $x$ onto the data manifold. Note that for  $x' \in M_{\cD_{train}}$, $J_{\cI}(x')$ has eigenvalues of $1$ or $0$. Therefore, the sign of the products of the non-zero eigenvalues will be $1$. Similarly, the arguments of the eigenvalues of $J_{\cI}(x')$ are $0$. Furthermore, as geodesics in Euclidean space are straight lines, any parallel transport of $x$ to $x'$, will not introduce rotation (including orientation reversal) of the of the basis of $T_x\cI$ . Therefore, the composition $f_{enc}: \cI \rightarrow \cL$ will not introduce rotation or orientation reversing behavior. 
\end{proof}

\newpage
\onecolumn
\section*{Addional figures}
\begin{table}
\begin{tabular}{cccc}
 & Bright ($\cD_{5,.8}$) & Moderate ($\cD_{normal}$) & Dark ($\cD_{.13}$) \\ 
$d =  2$ & \includegraphics[width=.3\textwidth]{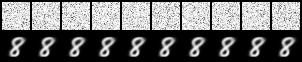} &\includegraphics[width=.3\textwidth]{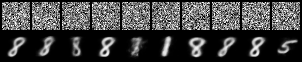} & \includegraphics[width=.3\textwidth]{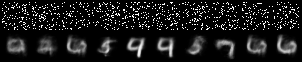} \\
$d =  3$ & \includegraphics[width=.3\textwidth]{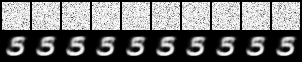} &\includegraphics[width=.3\textwidth]{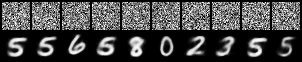} & \includegraphics[width=.3\textwidth]{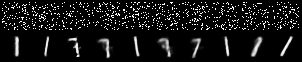} \\
$d =  4$ & \includegraphics[width=.3\textwidth]{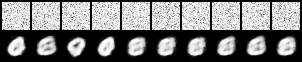} &\includegraphics[width=.3\textwidth]{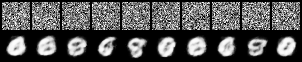} & \includegraphics[width=.3\textwidth]{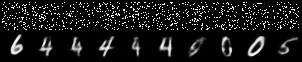} \\
$d =  5$ & \includegraphics[width=.3\textwidth]{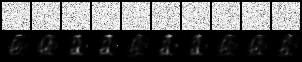} &\includegraphics[width=.3\textwidth]{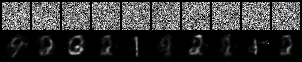} & \includegraphics[width=.3\textwidth]{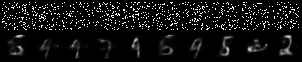} \\
$d =  6$ & \includegraphics[width=.3\textwidth]{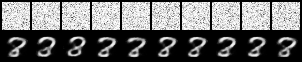} &\includegraphics[width=.3\textwidth]{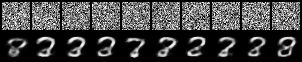} & \includegraphics[width=.3\textwidth]{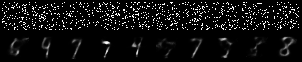} \\
$d =  7$ & \includegraphics[width=.3\textwidth]{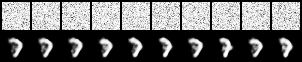} &\includegraphics[width=.3\textwidth]{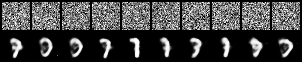} & \includegraphics[width=.3\textwidth]{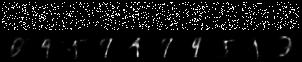} \\
$d =  8$ & \includegraphics[width=.3\textwidth]{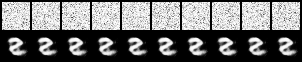} &\includegraphics[width=.3\textwidth]{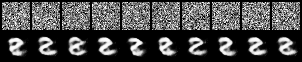} & \includegraphics[width=.3\textwidth]{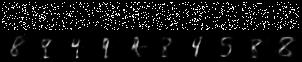} \\
$d =  9$ & \includegraphics[width=.3\textwidth]{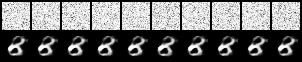} &\includegraphics[width=.3\textwidth]{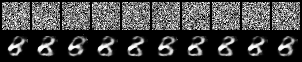} & \includegraphics[width=.3\textwidth]{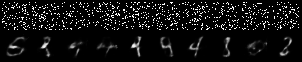} \\
$d =  10$ & \includegraphics[width=.3\textwidth]{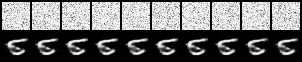} &\includegraphics[width=.3\textwidth]{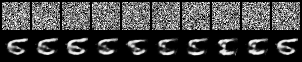} & \includegraphics[width=.3\textwidth]{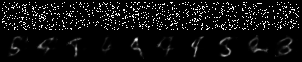} \\
$d =  11$ & \includegraphics[width=.3\textwidth]{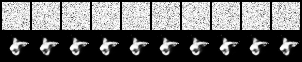} &\includegraphics[width=.3\textwidth]{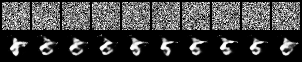} & \includegraphics[width=.3\textwidth]{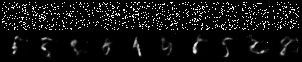} \\
$d =  12$ & \includegraphics[width=.3\textwidth]{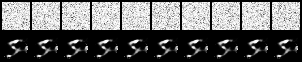} &\includegraphics[width=.3\textwidth]{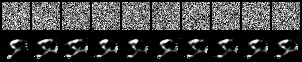} & \includegraphics[width=.3\textwidth]{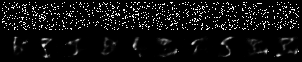} \\
$d =  13$ & \includegraphics[width=.3\textwidth]{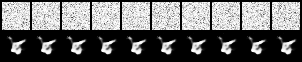} &\includegraphics[width=.3\textwidth]{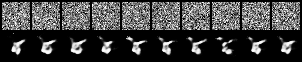} & \includegraphics[width=.3\textwidth]{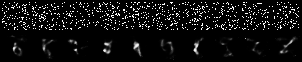} \\
$d =  14$ & \includegraphics[width=.3\textwidth]{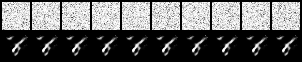} &\includegraphics[width=.3\textwidth]{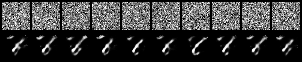} & \includegraphics[width=.3\textwidth]{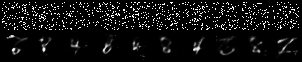} \\
$d =  15$ & \includegraphics[width=.3\textwidth]{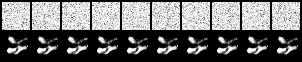} &\includegraphics[width=.3\textwidth]{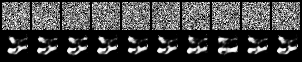} & \includegraphics[width=.3\textwidth]{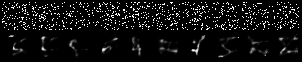} \\
$d =  16$ & \includegraphics[width=.3\textwidth]{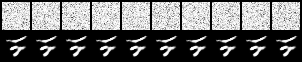} &\includegraphics[width=.3\textwidth]{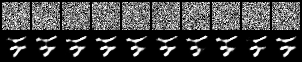} & \includegraphics[width=.3\textwidth]{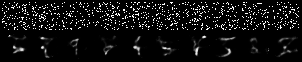} \\
$d =  17$ & \includegraphics[width=.3\textwidth]{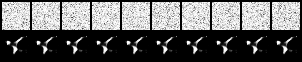} &\includegraphics[width=.3\textwidth]{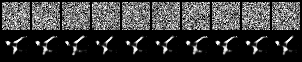} & \includegraphics[width=.3\textwidth]{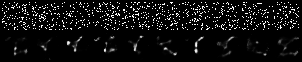} \\
$d =  18$ & \includegraphics[width=.3\textwidth]{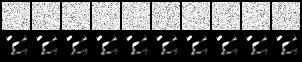} &\includegraphics[width=.3\textwidth]{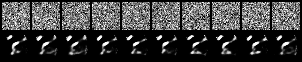} & \includegraphics[width=.3\textwidth]{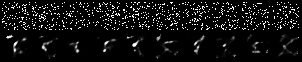} \\
$d =  19$ & \includegraphics[width=.3\textwidth]{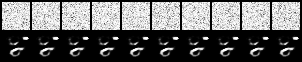} &\includegraphics[width=.3\textwidth]{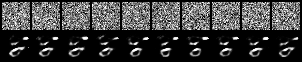} & \includegraphics[width=.3\textwidth]{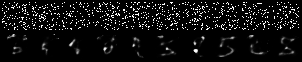} \\
$d =  20$ & \includegraphics[width=.3\textwidth]{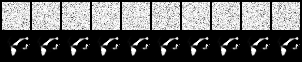} &\includegraphics[width=.3\textwidth]{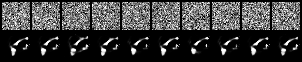} & \includegraphics[width=.3\textwidth]{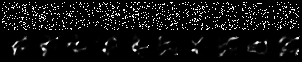} 
\end{tabular}
\caption{Reconstructions of a bright distribution ($\cD_{5, .8}$), and moderate meaned distribution ($\cD_{normal}$) and a dark distribution ($\cD_{.13}$) for seed 0. Note that by latent dimension 12, all of the reconstructioned images look like the same genearlized (non-MNIST) character. For low latent dimension, there is variation in the reconstructed images, which all look like MNIST characters. For moderate latent dimensions, the variations between classes has disappeared, but the reconstructions still look like MNIST characters.} 
\label{tbl:reconstructions}
\end{table}

\begin{table}
\begin{tabular}{cccc}
 & Bright ($\cD_{5,.8}$) & Moderate ($\cD_{normal}$) & Dark ($\cD_{.13}$) \\ 
$d =  2$ & \includegraphics[width=.3\textwidth]{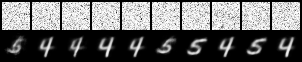} &\includegraphics[width=.3\textwidth]{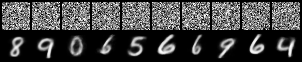} & \includegraphics[width=.3\textwidth]{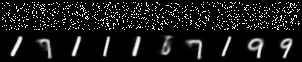} \\
$d =  3$ & \includegraphics[width=.3\textwidth]{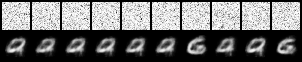} &\includegraphics[width=.3\textwidth]{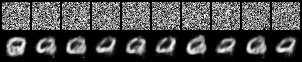} & \includegraphics[width=.3\textwidth]{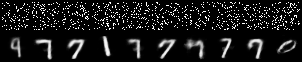} \\
$d =  4$ & \includegraphics[width=.3\textwidth]{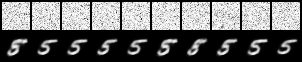} &\includegraphics[width=.3\textwidth]{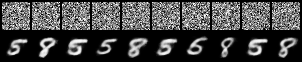} & \includegraphics[width=.3\textwidth]{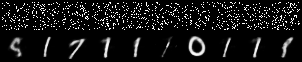} \\
$d =  5$ & \includegraphics[width=.3\textwidth]{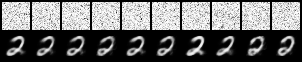} &\includegraphics[width=.3\textwidth]{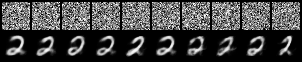} & \includegraphics[width=.3\textwidth]{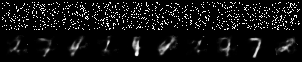} \\
$d =  6$ & \includegraphics[width=.3\textwidth]{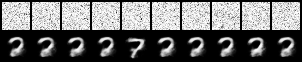} &\includegraphics[width=.3\textwidth]{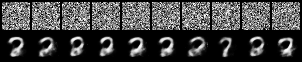} & \includegraphics[width=.3\textwidth]{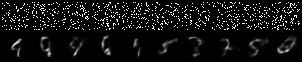} \\
$d =  7$ & \includegraphics[width=.3\textwidth]{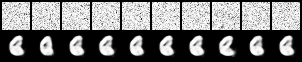} &\includegraphics[width=.3\textwidth]{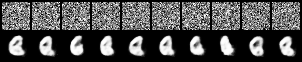} & \includegraphics[width=.3\textwidth]{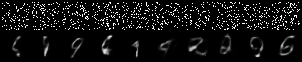} \\
$d =  8$ & \includegraphics[width=.3\textwidth]{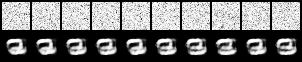} &\includegraphics[width=.3\textwidth]{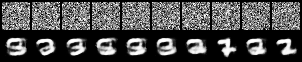} & \includegraphics[width=.3\textwidth]{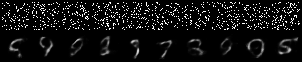} \\
$d =  9$ & \includegraphics[width=.3\textwidth]{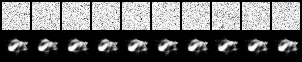} &\includegraphics[width=.3\textwidth]{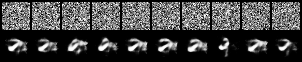} & \includegraphics[width=.3\textwidth]{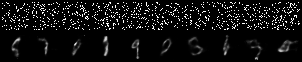} \\
$d =  10$ & \includegraphics[width=.3\textwidth]{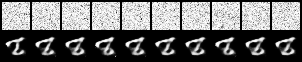} &\includegraphics[width=.3\textwidth]{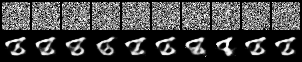} & \includegraphics[width=.3\textwidth]{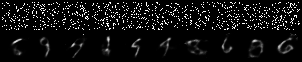} \\
$d =  11$ & \includegraphics[width=.3\textwidth]{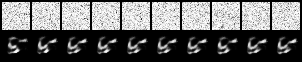} &\includegraphics[width=.3\textwidth]{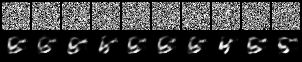} & \includegraphics[width=.3\textwidth]{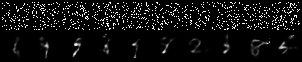} \\
$d =  12$ & \includegraphics[width=.3\textwidth]{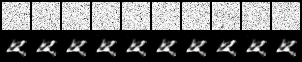} &\includegraphics[width=.3\textwidth]{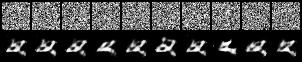} & \includegraphics[width=.3\textwidth]{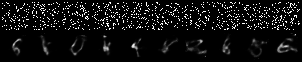} \\
$d =  13$ & \includegraphics[width=.3\textwidth]{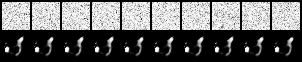} &\includegraphics[width=.3\textwidth]{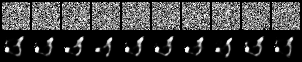} & \includegraphics[width=.3\textwidth]{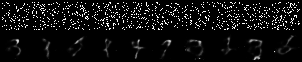} \\
$d =  14$ & \includegraphics[width=.3\textwidth]{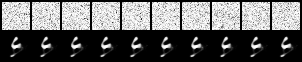} &\includegraphics[width=.3\textwidth]{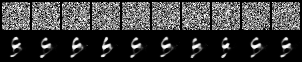} & \includegraphics[width=.3\textwidth]{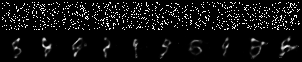} \\
$d =  15$ & \includegraphics[width=.3\textwidth]{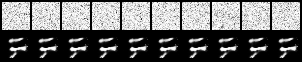} &\includegraphics[width=.3\textwidth]{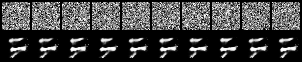} & \includegraphics[width=.3\textwidth]{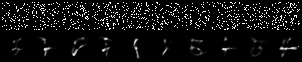} \\
$d =  16$ & \includegraphics[width=.3\textwidth]{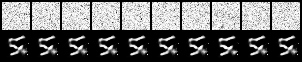} &\includegraphics[width=.3\textwidth]{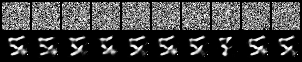} & \includegraphics[width=.3\textwidth]{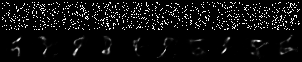} \\
$d =  17$ & \includegraphics[width=.3\textwidth]{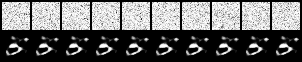} &\includegraphics[width=.3\textwidth]{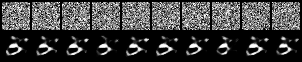} & \includegraphics[width=.3\textwidth]{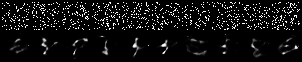} \\
$d =  18$ & \includegraphics[width=.3\textwidth]{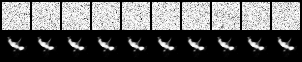} &\includegraphics[width=.3\textwidth]{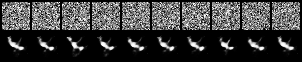} & \includegraphics[width=.3\textwidth]{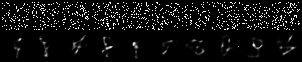} \\
$d =  19$ & \includegraphics[width=.3\textwidth]{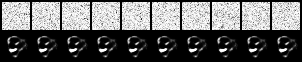} &\includegraphics[width=.3\textwidth]{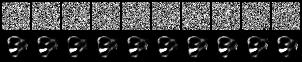} & \includegraphics[width=.3\textwidth]{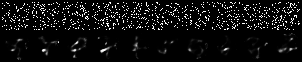} \\
$d =  20$ & \includegraphics[width=.3\textwidth]{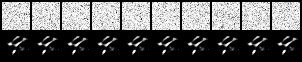} &\includegraphics[width=.3\textwidth]{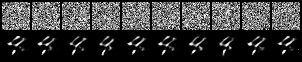} & \includegraphics[width=.3\textwidth]{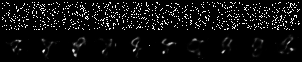} 
\end{tabular}
\caption{Reconstructions of a bright distribution ($\cD_{5, .8}$), and moderate meaned distribution ($\cD_{normal}$) and a dark distribution ($\cD_{.13}$) for seed 1. Note that for high latent dimension, all of the reconstructioned images look like the same genearlized (non-MNIST) character, but that these characters are different from those in seed 0 (shown in Figure \ref{tbl:reconstructions}.)} 
\label{tbl:reconstructions1}
\end{table}

\bibliographystyle{ieeetran}
\bibliography{The_OOD}

\end{document}